\documentclass[10pt,twocolumn,letterpaper]{article}

\usepackage[pagenumbers]{wacv} 

\newcommand{\cmark}{\ding{51}}%
\newcommand{\xmark}{\ding{55}}%

\def\vs{\emph{vs}\onedot}

\definecolor{wacvblue}{rgb}{0.21,0.49,0.74}
\usepackage[pagebackref,breaklinks,colorlinks,allcolors=wacvblue]{hyperref}
\usepackage{xcolor}
\usepackage{siunitx}
\usepackage{pifont}
\usepackage{csquotes} 
\usepackage{multirow}
\usepackage{tikz}

\title{AugMapNet: Improving Spatial Latent Structure via BEV Grid Augmentation for Enhanced Vectorized Online HD Map Construction}

\author{Thomas Monninger\textsuperscript{1,2} \hspace{0.3em} Md Zafar Anwar\textsuperscript{1} \hspace{0.3em} Stanislaw Antol\textsuperscript{1} \hspace{0.3em} Steffen Staab\textsuperscript{2,3} \hspace{0.3em} Sihao Ding\textsuperscript{1}
\vspace{3mm}
\\
\small{\textsuperscript{1}Mercedes-Benz Research \& Development North America, USA} \\
\small{\textsuperscript{2}University of Stuttgart, Germany} \\
\small{\textsuperscript{3}University of Southampton, United Kingdom}
}

\newcommand\copyrighttext{\footnotesize \textcopyright~2025 IEEE. Personal use of this material is permitted.  Permission from IEEE must be obtained for all other uses, in any current or future media, including reprinting/republishing this material for advertising or promotional purposes, creating new collective works, for resale or redistribution to servers or lists, or reuse of any copyrighted component of this work in other works.
}%

\newcommand\copyrightnotice{%
    \begin{tikzpicture}[remember picture,overlay]%
     \node[anchor=south, xshift=0pt, yshift=12pt] at (current page.south)%
     {\fbox{\parbox{\dimexpr\textwidth-\fboxsep-\fboxrule\relax}{\copyrighttext}}};%
     \end{tikzpicture}%
}

\begin{document}
\maketitle
\begin{abstract}    
Autonomous driving requires understanding infrastructure elements, such as lanes and crosswalks.
To navigate safely, this understanding must be derived from sensor data in real-time and needs to be represented in vectorized form.
Learned Bird's-Eye View (BEV) encoders are commonly used to combine a set of camera images from multiple views into one joint latent BEV grid.
Traditionally, from this latent space, an intermediate raster map is predicted, providing dense spatial supervision but requiring post-processing into the desired vectorized form.
More recent models directly derive infrastructure elements as polylines using vectorized map decoders, providing instance-level information.
Our approach, \textbf{Aug}mentation \textbf{Map Net}work (AugMapNet), proposes latent BEV feature grid augmentation, a novel technique that significantly enhances the latent BEV representation.
AugMapNet combines vector decoding and dense spatial supervision more effectively than existing architectures while remaining easy to integrate compared to other hybrid approaches.
It additionally benefits from extra processing on its latent BEV features.
Experiments on nuScenes and Argoverse2 datasets demonstrate significant improvements on vectorized map prediction of up to \SI{13.3}{\percent} over the StreamMapNet baseline on \SI{60}{\meter} range and greater improvements on larger ranges.
We confirm transferability by applying our method to another baseline, SQD-MapNet, and find similar improvements.
A detailed analysis of the latent BEV grid confirms a more structured latent space of AugMapNet and shows the value of our novel concept beyond pure performance improvement.
The code can be found at \href{https://github.com/tmonnin/augmapnet}{https://github.com/tmonnin/augmapnet}.
\end{abstract}

\vspace{-1mm}
\section{Introduction}
Autonomous driving relies on an accurate representation of the static infrastructure surrounding the autonomous vehicle to understand the environment and make informed decisions. 
Typically, a High-Definition (HD) map is given \textit{a priori} to capture essential information such as lanes, dividers, crosswalks, and other static elements.
However, the real world is subject to change due to events such as construction, road closures, or weather conditions.
Hence, the construction of online maps from the perception system is essential to ensure safety.
{\copyrightnotice}

\begin{figure}
    \centering
    \vspace{-12pt}
    \includegraphics[width=1\linewidth, trim=1.2cm 0.5cm 0.2cm 0.0cm, clip]{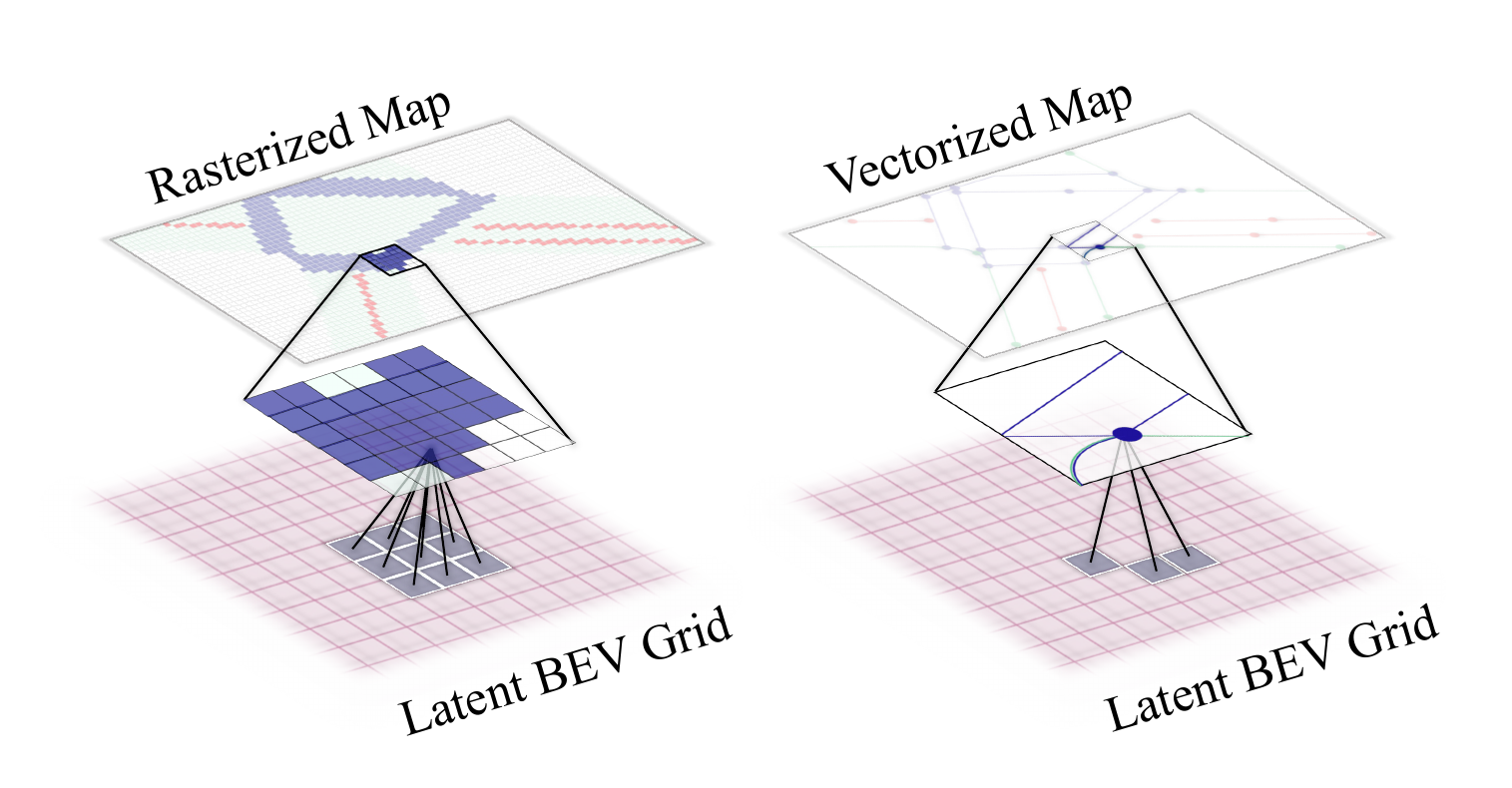}
    \vspace{-19pt}
    \caption{Schematic relation between output element and latent BEV grid. Rasterized decoding (left) and vectorized decoding (right) provide dense and sparse spatial supervision, respectively.}
    \label{fig:overview_augmapnet}
    \vspace{-10pt}
\end{figure}
An online HD map informs the autonomous vehicle about its environment in real time.
Maps represent characteristics of the traffic scene, where its static elements may not be fully defined, \eg, defining lanes purely based on the visible markings might fail in case of their absence.
Hence, characteristics of the traffic scene must be derived from the infrastructure context, which requires a holistic semantic understanding of the traffic scene. 
In the case of lanes, other infrastructure elements, such as curbs, must be considered.
Furthermore, most autonomous vehicle systems require the vectorized map for use in their planning systems, as it is computationally efficient and avoids ambiguity \cite{vectormapnet_2023, maptrv2_2023, gao2020vectornet}.

Related works for online HD map construction commonly perform pixel-level semantic segmentation in a 2D top-down view to predict a Bird's-Eye View (BEV) raster map \cite{pan2020cross, liftsplatshoot_2020, bevformer_2022}.
A vectorized representation can be derived from this intermediate raster map representation in a post-processing step \cite{hdmapnet_2022}.
Recent approaches \cite{vectormapnet_2023, maptrv2_2023, streammapnet_2024} introduce vector map decoding to predict the spatial position of static elements with polylines and thus directly generate the vectorized representation of the target.

Both paradigms leverage the same learned BEV encoder concept that fuses data from multiple sensors into one joint latent BEV feature grid \cite{liftsplatshoot_2020}, or BEV grid for shorthand, which is a spatial grid representation of the environment encoded in the latent feature space.
However, as shown in \cref{fig:overview_augmapnet}, rasterized map decoders (left) and vectorized map decoders (right) have different extents of spatial supervision.
First, raster map decoders predict pixel-level classifications.
Outputs and labels are both dense rasters, providing a dense pixel-wise loss for optimization.
In contrast, vector decoders have sparse outputs and labels, resulting in reduced guidance from the sparse loss.

Secondly, raster map decoders typically use convolutional layers, where each cell in the latent BEV grid contributes to the output.
In contrast, vector map decoders typically use deformable attention-based mechanisms, such as deformable DETR \cite{zhu2021deformable}, where only a subset of latent BEV grid cells contributes to the output.
Therefore, in a single training step, a raster map decoder optimizes all cells of latent BEV grid (dense spatial supervision), while vector map decoder optimizes only a subset (sparse supervision).

Based on this reasoning, we argue that including dense spatial supervision provides additional guidance for deriving the high-quality vectorized HD map.
A common approach is adding an auxiliary semantic segmentation task \cite{maptrv2_2023, pivotnet_2023, seabird_2024}. 
We propose a novel latent BEV feature grid augmentation method that directly injects dense spatial features from an intermediate predicted raster map into the latent BEV grid.
We further incorporate gradient stopping to treat the predicted raster map as an immutable prior.
To verify our method, we implement AugMapNet, a model that leverages latent BEV grid processing and augmentation to improve the performance of vector map decoders.

In summary, the main contributions of this paper are:
\begin{itemize}
\item We propose a latent BEV grid augmentation method that combines vector decoding and dense spatial supervision.
\item We develop a new model, AugMapNet, that adds latent augmentation for dense spatial supervision, along with gradient stopping and extra BEV processing, and demonstrate its effectiveness through extensive experiments.
\item We show that applying our method significantly enhances the structure in the latent BEV grid, which correlates with improved vector map decoding performance.
\end{itemize}

\section{Related Work}\label{sec:related_work}

\subsection{Rasterized Map Construction}
One option to represent map elements is a rasterized form (\cref{fig:overview_augmapnet}, left).
Philion and Fidler \cite{liftsplatshoot_2020} proposed the first learning-based architecture for online raster map construction from camera images by performing pixel-wise classification.
BEVFormer \cite{bevformer_2022} and subsequent works, including BEVFormer v2 \cite{bevformerv2_2023}, BEVDet4D \cite{bevdet4d_2022}, VideoBEV \cite{videobev_2024}, StreamPETR \cite{streampetr_2023}, SoloFusion \cite{park2022solofusion}, and TempBEV \cite{monninger2024tempbev}, improve accuracy by aggregating temporal information across multiple time steps.
Like previous approaches, HDMapNet \cite{hdmapnet_2022} performs semantic segmentation first, but adds a post-processing step to derive a vectorized representation.
While raster map prediction leads to dense spatial supervision, post-processing is required to get a vectorized form \cite{vectormapnet_2023, maptrv2_2023}, which is typically based on heuristics and restricts scalability and performance \cite{hdmapnet_2022, vectormapnet_2023}.
In AugMapNet, we directly predict vectorized representations and still utilize dense spatial supervision by using our method. 

\subsection{Vectorized Map Construction}
Recent approaches directly predict the vectorized representation of map elements.
Early works focus on topology and limit themselves to only lane centerlines \cite{hdmapgen_2021, zurn_lane_2021, can_structured_2021}.
VectorMapNet \cite{vectormapnet_2023} was the first end-to-end model to showcase the benefit of directly predicting a vectorized map representation including drivable areas, boundaries, dividers, and crosswalks.
MapTR \cite{maptrv2_2023} addresses the ambiguity in selecting a discrete set of points to model geometries in vectorized representations by employing permutation-equivalent modeling, which stabilizes the learning process.
StreamMapNet \cite{streammapnet_2024} uses memory buffers to achieve temporal stability, aiding in the construction of local HD maps.

All recent vector map decoders \cite{vectormapnet_2023, maptrv2_2023, streammapnet_2024, monninger2025mapdiffusion} share the same underlying deformable DETR \cite{zhu2021deformable} mechanism for decoding (see \cref{fig:overview_augmapnet}, right).
Its deformable attention mechanism samples $K$ keys as a subset of all the latent BEV grid feature vectors.
Only a subset of the BEV grid cells contributes to the sparse vectorized output, leading to sparse spatial supervision.
Hence, this type of vector map decoder is expected to exhibit a latent space with less spatial structure compared to raster map decoders.
In AugMapNet, we extend this type of existing models with a novel latent BEV grid augmentation method to combine the benefits of direct vectorized prediction and dense spatial supervision.

\subsection{Prior and Topology-Aware Map Methods}
External prior maps can enhance online HD map construction, a concept explored by works like P-MapNet \cite{pmapnet}, SMERF \cite{smerf_2024}, and NavMapFusion \cite{monninger2025navmapfusion}.
These methods leverage readily available Standard-Definition maps to improve performance, particularly at longer ranges.
In addition, SMERF and Chameleon \cite{zhang2025chameleon} predict complex topological structures, such as intersections and virtual lanes, usually overlooked by standard pipelines.
While these approaches rely on external prior data, our BEV grid augmentation is an orthogonal, prior-less method that uses a self-generated \enquote{prior} to enhance HD map construction.

\subsection{Latent Space Supervision and Analysis}
A straightforward approach to add spatial supervision is adding an auxiliary semantic segmentation task, as done by PivotNet \cite{pivotnet_2023}, MapTRv2 \cite{maptrv2_2023}, and SeaBird \cite{seabird_2024}.
MapVR \cite{zhang2024mapvr} adds a differentiable rasterizer on top of a vector map decoder only for training.
A downstream auxiliary segmentation loss provides supervision directly to the vectorized output.
We extend their argument that more sophisticated techniques can increase performance beyond pure auxiliary supervision \cite{zhang2024mapvr} by introducing the AugMapNet approach, where the segmentation loss does not backpropagate to the latent BEV grid and is therefore not an auxiliary loss.
 
Two recent works also leverage a raster representation beyond auxiliary supervision.
HIMap \cite{zhou2024himap} introduces a hybrid representation that jointly decodes both raster and vector outputs. 
In contrast, we adopt a two-stage approach in which raster decoding preconditions subsequent vector decoding.
MGMap \cite{liu2024mgmap} likewise employs a two-stage strategy, generating an intermediate rasterized representation from which the vectorized map is predicted. 
However, in two-stage designs, the raster output can become a high-frequency latent space \cite{chu2017cyclegan}. 
To address this, we apply gradient stopping to treat the intermediate representation as an independent prior.
In summary, our mechanism combines raster and vector representations better than other hybrid architectures, while being modular enough to be easily transferred to different models.
It augments the latent BEV space and preconditions the vector map decoding, which adds more structure to the final latent BEV grid.

We support our claims via analysis and visualizations on the latent BEV grid features with PCA \cite{hotelling1933pca}.
Since our evaluation already demonstrates that learning a non-linear decoder on these latent features results in superior metric scores, we opt for the simplicity of PCA instead of non-linear approaches, such as t-SNE \cite{van2008tsne} and UMAP \cite{mcinnes2018umap}.

Other works in related tasks leverage latent BEV grid visualizations, but often do not analyze the latent BEV grid for further insights into their mechanisms.
SA-BEV \cite{zhang2023sabev}, which proposes a pooling mechanism to improve the semantics of the latent BEV grid features, and BEV-SAN \cite{chi2023bevsan}, which adds height slices to the BEV grid, only visualize each cell's feature vector norm to produce a grayscale image without additional quantitative analysis.
FB-BEV \cite{li2023fbbev} uses RGB BEV feature visualizations to support its claims, but lacks details on how BEV features are reduced to RGB and also lacks analysis of the BEV features.

Instead of cameras, some works that visualize and analyze latent features use LiDAR input.
Range image visualizations rely on t-SNE in RangeRCNN \cite{liang2020rangercnn}, whereas others \cite{perez2024class, saltori2023lidog} use t-SNE to show class-colored latent features that form distinct clusters.
Perez \etal \cite{perez2024class} analyze their BEV features via the average relative distances between classes, whereas we use the mean Silhouette score \cite{rousseeuw1987silhouettes}.

\section{Approach}

\begin{figure*}
    \centering
    \includegraphics[width=1.0\textwidth]{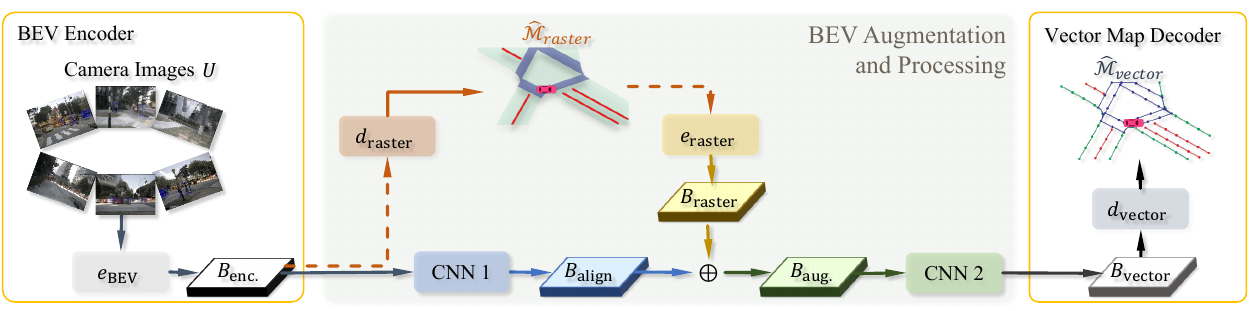}
    \caption{Overview of AugMapNet architecture. Camera images $U$ are processed by learned BEV encoder $e_{\mathrm{BEV}}$ into latent BEV grid $B_{\mathrm{enc}}$. AugMapNet employs a latent BEV grid augmentation mechanism that generates $\hat{M}_{\mathrm{raster}}$ (\ie, map semantic segmentation). Additional CNNs help structure the latent space. A vector map decoder $d_{\mathrm{vector}}$ predicts vectorized map $\hat{M}_{\mathrm{vector}}$. Dashed lines indicate gradient stop.}
    \label{fig:augmapnet_architecture}
\end{figure*}

\subsection{Problem Statement} \label{sec:problem}
Let $U=\{u_1, \ldots, u_n\}$ be the set of image frames from the $n$ monocular cameras mounted on the autonomous vehicle.
Let $\mathcal{M}_{\mathrm{vector}} = \{P_{\mathrm{div}}, P_{\mathrm{bound}}, P_{\mathrm{ped}}\}$ be the local HD map with the autonomous vehicle at the origin, where $P_{\mathrm{div}}$, $P_{\mathrm{bound}}$, and $P_{\mathrm{ped}}$ are the sets of polylines representing lane dividers, road boundaries, and pedestrian crossings within the scene, respectively.
Each polyline $P_k = \left[ (x_{i},y_{i} ) \right]_{i=1}^{N_{P_{k}}}$ is a sequence of ${N_{P_{k}}}$ points.
The goal is to find a function $f$ that predicts the local vectorized HD map $\mathcal{\hat{M}}_{\mathrm{vector}}$ from the set of image frames ${U}$, \ie, $\mathcal{\hat{M}}_{\mathrm{vector}} = f \left( U \right)$.
The function $f$ is typically realized in an encoder-decoder fashion \cite{streammapnet_2024,vectormapnet_2023,maptrv2_2023,hdmapnet_2022}.
An encoder $e_{\mathrm{BEV}}$ generates the latent BEV grid $B$, \ie, $B = e_{\mathrm{BEV}}({U})$.
A decoder $d_{\mathrm{vector}}$ performs vector decoding into the local HD map representation, \ie, $\mathcal{\hat{M}}_{\mathrm{vector}} = d_{\mathrm{vector}} (B)$.

\subsection{Latent BEV Grid Augmentation} \label{seg:bev_space_augmentation}
First, we provide a mathematical framework for the terms \enquote{dense} and \enquote{sparse} spatial supervision.
Generally, the loss is calculated over all elements of the output representation $\hat{\mathcal{M}}$.
Supervision for a step is done by optimizing learnable weights $W$ based on the gradients of the loss:
$\frac{\partial L}{\partial W} = \frac{\partial L}{\partial \hat{\mathcal{M}}}  \frac{\partial \hat{\mathcal{M}}}{\partial W}$.
The total amount of spatial supervision therefore is related to the number of elements in $\hat{\mathcal{M}}$.
In the case of a raster map decoder, the total gradient signal is calculated from the dense loss over all cells of the raster map ($\mathbb{R}^{H \times W}$), which gives the total number of elements $| \hat{\mathcal{M}}_{\mathrm{raster}} | = H \cdot W$.
In the case of a vector map decoder, the total gradient signal is calculated from the sparse loss over $N_p$ points of $N$ polylines.
The number of elements providing supervision is at most $| \hat{\mathcal{M}}_{\mathrm{vector}} | = N \cdot N_p$.
In typical settings, such as AugMapNet, this inequality on cardinality generally holds: $|\hat{\mathcal{M}}_{\mathrm{raster}}| > |\hat{\mathcal{M}}_{\mathrm{vector}}|$.
This quantifies the benefit of dense spatial supervision (see more details in Supplementary \ref{sec:spatial_formulation_supplementary}).
Note that, in DETR-style decoders, out of $N$ predicted elements, the ones correctly predicted as empty class do not induce spatial supervision.
Hence, this number is the upper bound and the actual supervision is typically much sparser in a single step.

Given this motivation for dense spatial supervision, the goal is to combine rasterized prediction with decoding a vector map representation.
To this end, we propose a novel concept of augmenting the latent BEV grid $B$ with additional information from a predicted raster map.
During training, we rasterize the ground truth $\mathcal{M}_{\mathrm{vector}}$ into $\mathcal{M}_{\mathrm{raster}}$ and train a parallel raster map decoder to predict $\mathcal{\hat{M}}_{\mathrm{raster}}$.
Given an encoder $e_{\mathrm{raster}}$ that preserves the spatial structure, the BEV augmentation mechanism is defined as:
\begin{equation}
    B_{\mathrm{aug}} = B + e_{\mathrm{raster}}(\mathcal{\hat{M}}_{\mathrm{raster}}).
    \label{eq:bev_augmentation}
\end{equation}
By injecting the spatial features into the latent BEV grid, we expect to enhance its spatial structure.
Directly adding the encoded spatial features is expected to have a stronger impact than auxiliary supervision.
$B_{\mathrm{aug}}$ is then used instead of $B$ in the vector map decoder to produce the final prediction.

\subsection{AugMapNet Model Architecture} \label{sec:model}
We design a novel model, AugMapNet, to demonstrate the benefit of our proposed latent BEV grid augmentation concept.
The full architecture is shown in \cref{fig:augmapnet_architecture}.
On a high level, AugMapNet extends any given encoder-decoder structure with $e_{\mathrm{BEV}}$ and $d_{\mathrm{vector}}$ with a BEV space augmentation.
AugMapNet uses a learned BEV encoder $e_{\mathrm{BEV}}$ to encode given camera images $U$ into the BEV grid $B_{\mathrm{enc}}$. 
We add a parallel rasterized map decoder $d_{\mathrm{raster}}$ that predicts the rasterized map $\mathcal{\hat{M}}_{\mathrm{raster}}$ from $B_{\mathrm{enc}}$.
Following the BEV augmentation concept from \cref{eq:bev_augmentation}, $\mathcal{\hat{M}}_{\mathrm{raster}}$ is encoded by $e_{\mathrm{raster}}$ and added back to the latent BEV grid $B_{\mathrm{enc}}$, resulting in the augmented BEV grid $B_{\mathrm{aug}}$.
We introduce additional BEV processing with 2 CNN blocks to better leverage the latent BEV grid.
CNN1 transforms $B_{\mathrm{enc}}$ into an aligned BEV grid, $B_{\mathrm{align}}$, before it is augmented by $B_{\mathrm{raster}}$.
CNN2 operates on $B_{\mathrm{aug}}$ to combine the summed features into new latent space features in $B_{\mathrm{align}}$.
The benefit of these CNNs is demonstrated in \cref{sec:ablation_bev_augmentation}, where ablations are done without the CNNs.
The final latent BEV grid representation, $B_{\mathrm{vector}}$, is passed to the vector map decoder $d_{\mathrm{vector}}$ to extract the vectorized map representation $\mathcal{\hat{M}}_{\mathrm{vector}}$, comprising polylines $\hat{P}_{\mathrm{div}}$, $\hat{P}_{\mathrm{bound}}$, and $\hat{P}_{\mathrm{ped}}$.

\subsection{Loss and Gradient Stopping}\label{sec:loss}
The vector map decoder loss ($\mathcal{L}_\mathrm{vector}$) is the same as in StreamMapNet \cite{streammapnet_2024} and consists of a polyline matching loss ($\mathcal{L}_\mathrm{line}$), a focal loss ($\mathcal{L}_\mathrm{focal}$), and a transformation loss ($\mathcal{L}_\mathrm{trans}$).
For the raster map decoder loss ($\mathcal{L}_{\mathrm{raster}}$), we evaluate semantic segmentation using the Dice loss \cite{milletari2016dice}.
Supplementary \ref{sec:loss_supplementary} has the mathematical definition of all losses.

Optimizing a parallel raster representation similar to MGMap \cite{liu2024mgmap} poses the risk of creating a high-frequency latent space due to the superimposed vector map decoder loss, which can hinder interpretability (\cf \cite{chu2017cyclegan}). 
To mitigate this issue, our latent BEV grid augmentation mechanism utilizes gradient stopping, as illustrated by the dashed lines in \cref{fig:augmapnet_architecture}.
This way, the vector decoding task is decoupled from the semantic segmentation task.
Specifically, gradients are not backpropagated from $d_{\mathrm{raster}}$ to $e_{\mathrm{BEV}}$ to ensure that the semantic segmentation task does not affect the latent BEV grid $B_{\mathrm{enc}}$ and hence does not compromise the vector map decoding performance.
Furthermore, gradients are not backpropagated from $e_{\mathrm{raster}}$ to $d_{\mathrm{raster}}$ to avoid $d_{\mathrm{raster}}$ using the rasterized map prediction $\hat{M}_{\mathrm{raster}}$ as a high-frequency latent space for vector map features.
The vector decoder treats the predicted rasterized map as an immutable prior and $e_{\mathrm{raster}}$ learns how to generate features that are useful for the vector map decoding task.
Different variations of gradient stopping are evaluated in \cref{sec:ablation_bev_augmentation}.

\subsection{Structure of Latent BEV Grid}\label{sec:latent_structure}
We hypothesize that our latent BEV augmentation induces a more structured latent space due to dense spatial supervision.
To measure the structure, we first use cluster analysis on the latent BEV grid on which $d_{\mathrm{vector}}$ operates (\ie, $B_{\mathrm{vector}}$).
Class indices from the ground truth rasterized map are used as cluster labels for each of the latent vectors (\ie, each BEV grid cell).
The Silhouette score \cite{rousseeuw1987silhouettes} is used to quantify cluster cohesion and separation for cluster quality evaluation and we use its mean of all data points:

\begin{equation}
    S = \frac{1}{N} \sum_{i=1}^{N} \frac{b(i) - a(i)}{\max(a(i), b(i))},
    \label{eq:silhouette_score}
\end{equation}
where \( a(i) \) is the average distance between \( i \) and all other points in its cluster and \( b(i) \) is the minimum average distance from \( i \) to the nearest cluster that \( i \) is not a part of.

Secondly, we leverage a dimensionality reduction technique that linearly transforms the high-dimensional latent space into a low-dimensional space that can be visualized and studied.
Specifically, we perform a Principal Component Analysis (PCA) \cite{hotelling1933pca} on each scene's latent features and analyze the top 3 components ordered by the amount of variance they capture from the data.
We quantify the similarity between GT label and latent BEV grid for each scene by measuring the Mutual Information \cite{shannon1948mathematical} between the ground truth (GT) raster map $\mathcal{M}_{\mathrm{raster}}$ and the top 3 principal components of $B_{\mathrm{vector}}$.
Formally, given two sets of labels $U$ and $V$, Mutual Information is defined as :
\begin{equation}
    MI(U, V) = \sum_{i=1}^{|U|} \sum_{j=1}^{|V|} \frac{|U_i \cap V_j|}{N} \log \left( \frac{N |U_i \cap V_j|}{|U_i| |V_j|} \right).
    \label{eq:mutual_information}
\end{equation}
In our case, $U$ represents the flattened ground truth raster map after representing each cell with a unique class index.
Following common practice for mutual information estimation \cite{kraskov2004estimating}, we partition the continuous space of the PCA result into bins of finite size and choose 10 equidistant bins per dimension.
$V$ represents the flattened PCA result after binning and mapping each cell to a unique index.

\section{Experiments} \label{sec:experiments}

\subsection{Datasets and Evaluation Metrics}\label{sec:datasets_metrics}
We conduct our experiments on nuScenes \cite{nuscenes}, which provides samples at \SI{2}{\hertz}, and Argoverse2 \cite{argoverse}, which we resample to \SI{2}{\hertz} for consistency. 
We use images from the 6 (nuScenes) or 7 (Argoverse2) monocular cameras and corresponding ground truth vectorized map elements: road boundary ($\mathrm{bound})$, lane dividers ($\mathrm{div}$), and pedestrian crossings ($\mathrm{ped}$).
We use StreamMapNet's \cite{streammapnet_2024} train and validation splits with no geospatial overlap for evaluation.

The performance on the vector map decoding task is evaluated for each of the three classes of polylines using Average Precision (AP) and combined with mean Average Precision (mAP).
Following StreamMapNet, the AP thresholds are \{0.5, 1.0, 1.5\} and \{1.0, 1.5, 2.0\} for the $\SI{60}{\meter} \times \SI{30}{\meter}$ and $\SI{100}{\meter} \times \SI{50}{\meter}$ perception ranges, respectively.
Intersection over Union (IoU) is evaluated per class for the raster map decoding task and reported as mean IoU (mIoU).
The model runtime is evaluated in frames per second (FPS).

\subsection{Baseline Models}\label{sec:baseline}
We use several state-of-the-art models for vectorized map prediction as baselines: MapTR \cite{maptrv2_2023}, P-MapNet \cite{pmapnet}, SQD-MapNet \cite{wang2024stream}, and StreamMapNet \cite{streammapnet_2024}.
Additionally, we compare against hybrid methods that combine raster and vector representations: MapTRv2 \cite{maptrv2_2023}, MapVR \cite{zhang2024mapvr}, and MGMap \cite{liu2024mgmap}. 
For a fair and direct comparison, we ran all baselines on nuScenes and method-specific baselines on Argoverse2 ourselves and used the settings from StreamMapNet and SQD-MapNet for training AugMapNet and AugMapNet-SQD (AugSQD for short), respectively.

\subsection{Implementation Details}\label{sec:implementation}
StreamMapNet \cite{streammapnet_2024} is our baseline implementation with BEVFormer \cite{bevformer_2022} as the BEV encoder $e_{\mathrm{BEV}}$ and temporal aggregation is done with a Gated Recurrent Unit (GRU) \cite{cho_2014_gru}.
The vector map decoder $d_{\mathrm{vector}}$ is the custom deformable DETR decoder \cite{zhu2021deformable} from StreamMapNet \cite{streammapnet_2024} that uses Multi-Point Attention.

We perform semantic segmentation on the three classes: lane dividers, road boundaries, and pedestrian crossings.
Thus, $\mathcal{\hat{M}}_{\mathrm{raster}} \in \mathbb{R}^{100 \times 50 \times 3}$.
The decoder $d_{\mathrm{raster}}$ is a 4-layer CNN with output channels $\{128, 64, 32, 3\}$.
The encoder $e_{\mathrm{raster}}$ is a 4-layer CNN with output channels $\{32, 64, 128, 256\}$.
CNN1 and CNN2 both have 2 layers with input and output channel sizes of 256.
All CNNs use convolutions with kernel size 3 and stride 1, batch normalization, and ReLU.
With this configuration, AugMapNet has 52.70 million parameters, \SI{6.3}{\percent} more than StreamMapNet.
We train the model for 24 epochs using a batch size of 1, with the training performed in parallel on 8 Nvidia V100 GPUs.
AdamW optimizer is used with cosine annealing and a $2.5 \times 10^{-4}$ learning rate.

\subsection{Quantitative Results}\label{sec:quantitative_results}

\cref{tab:table_main_result} shows the results of our AugMapNet model in comparison to relevant baseline models on the nuScenes dataset.
Naive auxiliary semantic segmentation ($+ \mathrm{Aux.}$) by adding $d_{\mathrm{raster}}$ optimized with $\mathcal{L}_{\mathrm{raster}}$  gives a slight improvement with \SI{34.0}{\percent}~mAP.
The results show that AugMapNet outperforms all baselines on the overall mAP, and on each class-specific metric except for P-MapNet on AP$_{\mathrm{bound}}$ due to its map prior.
Overall, AugMapNet achieves \SI{38.3}{\percent} mAP \vs \SI{33.8}{\percent} mAP for StreamMapNet.
This translates to \SI{+13.3}{\percent} relative improvement over the state-of-the-art StreamMapNet model.
While the relative improvements on lane dividers (\SI{+11.0}{\percent}) and road boundaries (\SI{+5.6}{\percent}) are strong, the relative improvement on pedestrian crossings is immense (\SI{+26.3}{\percent}).
The FPS of AugMapNet differs only slightly from StreamMapNet with a relative change of \SI{-9.6}{\percent}.
The model size also remains almost the same at \SI{1.8}{\giga\byte} for both StreamMapNet and AugMapNet.
During training with batch size 1, we observe \SI{\approx 7.7}{\giga\byte} VRAM usage, a \SI{+1.6}{\percent} increase over StreamMapNet.
More details are given in Supplementary~\ref{sec:non-functional}.

\cref{tab:ablation_perception_ranges} shows results on a larger perception range and on applying our method to a different model.
Beyond the reported \SI{+13.3}{\percent} improvement on $\SI{60}{\meter}$, we get \SI{+24.6}{\percent} on $\SI{100}{\meter}$. 
We provide further ablation on perception ranges in Supplementary \ref{sec:ablation_range}, showing an increase in relative improvement at larger perception ranges.
To verify generalizability of our method, we apply our proposed latent BEV augmentation method to another model, SQD-MapNet \cite{wang2024stream}, and denote it as AugSQD.
The improvement is \SI{+9.8}{\percent} on perception range $\SI{60}{\meter}$ and \SI{+21.1}{\percent} on the $\SI{100}{\meter}$ range.

To assess the generalization of our method across datasets and sensor configurations, we trained AugMapNet on Argoverse2 dataset \cite{argoverse} without any adjustments.
Argoverse2 covers different cities and a different sensor configuration (7 \vs 6 cameras).
\cref{tab:table_argo_result} shows the results of our AugMapNet model in comparison to relevant baseline models.
Again, AugMapNet outperforms all other approaches and reaches \SI{58.8}{\percent} mAP, a significant improvement over StreamMapNet, confirming the broader applicability of our method. 
Results on Argoverse2 for larger perception ranges and AugSQD are given in Supplementary~\ref{sec:results_augmapnet-sqd_argoverse2}.

{\begin{table}[tbp]
    \centering
    \resizebox{0.95\columnwidth}{!}{%
    \begin{tabular}{l|cccc|c}
        Method & AP$_{\mathrm{ped}}$&  AP$_{\mathrm{div}}$  &  AP$_{\mathrm{bound}}$& mAP & FPS   \\ \midrule
        MapTR \cite{maptr_2023}       & 7.5 & 23.0 & 35.8 & 22.1 & \textbf{16.0}\\
        MapVR$^\dagger$ \cite{zhang2024mapvr}       & 10.1 & 22.6 & 35.7 & 22.8 & \textbf{16.0} \\
        MGMap$^\dagger$ \cite{liu2024mgmap}       & 7.9 & 25.6 & 37.4 & 23.7 & 11.8 \\
        MapTRv2$^\dagger$ \cite{maptrv2_2023}       & 16.2 & 28.7 & 44.8 & 29.9 & 14.9 \\
        P-MapNet$^\star$\cite{pmapnet}& 17.7 & 26.1 & \textbf{48.4} & 30.7 & 12.2\\
        SQD-MapNet \cite{wang2024stream}  & 34.8& 31.7& 44.3& 36.9& 11.9 \\
        StreamMapNet \cite{streammapnet_2024} &  31.2 & 27.3 & 42.9 & 33.8 & 13.2 \\
        StreamMapNet + Aux.$^\dagger$    & 31.9 & 27.7 & 42.3 & 34.0 & 13.2\\
        AugMapNet$^\dagger$ (ours) & 39.4 &  30.3 &  45.3 & 38.3 & 11.9 \\
        AugSQD$^\dagger$ (ours) & \textbf{41.1} & \textbf{32.6} & 47.8 & \textbf{40.5} & 10.8 \\
    \end{tabular}%
    }
    \caption{AugMapNet compared to baselines at perception range $\SI{60}
    {\meter} \times \SI{30}{\meter}$ on nuScenes split without geospatial overlap \cite{roddick2020pyroccnet}. 
    \enquote{AugSQD} is the application of our method to SQD-MapNet \cite{wang2024stream}. 
    $^\star$ denote the use of prior map, results from \cite{pmapnet}. $^\dagger$ denotes hybrid architectures. \enquote{Aux.} is auxiliary segmentation (see \Cref{sec:quantitative_results}).}
    \label{tab:table_main_result}
\end{table}
}
{\begin{table}[tbp]
    \centering
    \resizebox{0.9\columnwidth}{!}{%
    \begin{tabular}{p{3.1cm}|cccc|c}
        Method & AP$_{\mathrm{ped}}$ & AP$_{\mathrm{div}}$ & AP$_{\mathrm{bound}}$ & mAP & FPS   \\ \midrule
        VectorMapNet \cite{vectormapnet_2023}     & 35.6 & 34.9 & 37.8 & 36.1 & 5.5 \\
        MapTR \cite{maptr_2023}       & 48.1 & 50.4 & 55.0 & 51.1 & \textbf{18.0}\\
        StreamMapNet \cite{streammapnet_2024}  &  56.0 & 54.4 & 61.0 & 57.1 & 14.2\\
        AugMapNet (ours) & \textbf{57.4} & \textbf{57.4} & \textbf{61.6} & \textbf{58.8} & 12.8 \\
    \end{tabular}%
    }
    \caption{Performance on Argoverse2. Baseline results from \cite{streammapnet_2024}.}
    \label{tab:table_argo_result}
    \vspace{-6pt}
\end{table}
}

\subsection{Qualitative Results}\label{sec:qualitative_results}
\begin{figure}
    \centering
    \vspace{3pt}
    \includegraphics[width=\columnwidth]{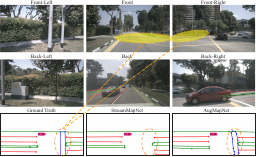}
    \vspace{-6mm}
    \caption{Qualitative result on nuScenes dataset. Input camera images ($U$) are at the top. Ground truth labels ($\mathcal{M}_{\mathrm{vector}}$) and predicted vectorized maps ($\hat{\mathcal{M}}_{\mathrm{vector}}$) are at the bottom. The orange circle highlights the crosswalk missed by StreamMapNet.}
    \label{fig:qualitative_result}
    \vspace{-8pt}
\end{figure}

\cref{fig:qualitative_result} shows qualitative results of StreamMapNet and AugMapNet on a traffic scene from the nuScenes dataset.
Both models correctly predict the lane dividers and road boundaries.
AugMapNet detects the crosswalk in front of the autonomous vehicle that is missed by StreamMapNet, which exemplifies our large gains on pedestrian crossings.
More examples can be found in the Supplementary \ref{sec:qualitative_results_extended}.

\subsection{Results on Latent BEV Grid Analysis}
\begin{figure}
    \centering
    \includegraphics[width=0.98\columnwidth]{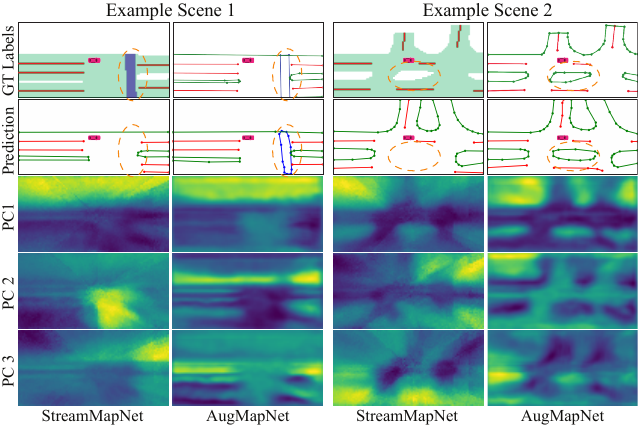}
    \vspace{-4pt}
    \caption{Rendering of GT $\mathcal{M}_{\mathrm{raster}}$ and $\mathcal{M}_{\mathrm{vector}}$, prediction $\hat{\mathcal{M}}_{\mathrm{vector}}$, and the top 3 principal components of the latent BEV grid input to $d_{\mathrm{vector}}$ for StreamMapNet and AugMapNet (PC1-3). Orange circles highlight elements missed by StreamMapNet.}
    \label{fig:latent_pca}
    \vspace{-4pt}
\end{figure}

We compare the latent BEV grid features of StreamMapNet and AugMapNet with respect to: 1) cluster analysis, 2) similarity to GT label, 3) variance.
First, we use cluster analysis on the latent BEV grid input to $d_{\mathrm{vector}}$ (\ie, $B_{\mathrm{vector}}$ in AugMapNet) to verify our hypothesis that latent BEV grid augmentation increases structure of the latent BEV grid. 
For scene 2 in \cref{fig:latent_pca}, AugMapNet reaches a Silhouette score of 0.119 \vs 0.086 for StreamMapNet.
Over the full nuScenes val dataset, the mean Silhouette score for AugMapNet is 4.5 times higher than StreamMapNet (0.076 \vs 0.017).
Therefore, features from the BEV grid of the same class are better clustered in the latent space of AugMapNet compared to StreamMapNet, which could benefit classification.
Since the ground truth labels have spatial structure, feature consistency among neighboring cells will also give a more consistent signal to the decoder.

{\begin{table}
    \centering
    \setlength{\tabcolsep}{4pt}
    \resizebox{\columnwidth}{!}{%
    \begin{tabular}{p{2cm}l|cccc|c}
        Range & Variant & AP$_{\mathrm{ped}}$ &  AP$_{\mathrm{div}}$ &  AP$_{\mathrm{bound}}$ & mAP & Impr. \\ \midrule
        \multirow{2}{*}{$\SI{60}{\meter}\!\times\!\SI{30}{\meter}^\ddagger$} & StreamMapNet \cite{streammapnet_2024} &  31.2 & 27.3 & 42.9 & 33.8 &  \\
        & AugMapNet (ours)  & 39.4 & 30.3 & 45.3 & 38.3 & \SI{+13.3}{\percent} \\ \midrule
        \multirow{2}{*}{$\SI{100}{\meter}\!\times\!\SI{50}{\meter}^\ast$} & StreamMapNet \cite{streammapnet_2024} & 25.5 & 19.3  & 24.7  & 23.2 &  \\
        & AugMapNet (ours) & 35.5 & 22.8  & 28.4  & 28.9 & \SI{+24.6}{\percent}\\ \midrule
        \multirow{2}{*}{$\SI{60}{\meter}\!\times\!\SI{30}{\meter}^\ddagger$} & SQD-MapNet \cite{wang2024stream} & 34.8& 31.7& 44.3& 36.9& \\
        & AugSQD (ours) & 41.1& 32.6& 47.8& 40.5& \SI{+9.8}{\percent}\\ \midrule
        \multirow{2}{*}{$\SI{100}{\meter}\!\times\!\SI{50}{\meter}^\ast$} & SQD-MapNet \cite{wang2024stream} & 26.9& 20.9& 27.6& 25.1& \\
        & AugSQD (ours) & 33.0& 24.7& 33.6& 30.4& \SI{+21.1}{\percent}\\ 
    \end{tabular}%
    }
    \caption{Results on different perception ranges for AugMapNet and AugSQD. Results on nuScenes \cite{nuscenes}. AP thresholds $^\ddagger$: \{0.5, 1.0, 1.5\}, $^\ast$: \{1.0, 1.5, 2.0\}.}
    \label{tab:ablation_perception_ranges}
    \vspace{-12pt}
\end{table}
}

\begin{figure*}
    \centering
    \begin{subfigure}[t]{0.49\textwidth}
        \centering
        \includegraphics[trim=0cm 0cm 0cm 0cm, clip, width=\textwidth]{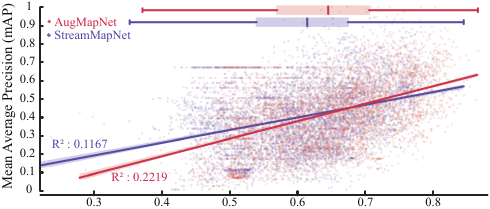}
        \caption{Mutual Information}
        \label{fig:mutual_information}
    \end{subfigure}
    \hfill
    \begin{subfigure}[t]{0.49\textwidth}
        \centering
        \includegraphics[trim=0cm 0cm 0cm 0cm, clip, width=\textwidth]{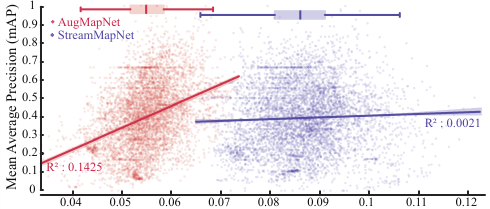}
        \caption{Variance of Latent BEV Grid}
        \label{fig:variance_analysis}
    \end{subfigure}
   \vspace{-5pt}
   \caption{Vectorized map prediction performance versus (a) Mutual Information between top 3 principal components and ground truth raster map and (b) Variance of latent BEV grid for all nuScenes val data points. (StreamMapNet: blue, AugMapNet: red)}
   \vspace{-2pt}
\end{figure*}

Secondly, to assess the similarity between GT label and latent BEV grid, we perform a PCA for each latent space and visualize the top 3 principal components ordered by the amount of captured variance for StreamMapNet and AugMapNet on two nuScenes data points in \cref{fig:latent_pca} (more results are in Supplementary \ref{sec:pca_plots}).
The results show spatial similarity with both the vectorized and rasterized map ground truths.
Areas like background and drivable roads are well segmented and retrievable from the principal components.
In both scenes, similarity appears higher for AugMapNet, which matches the cluster analysis results.
AugMapNet results also have sharper boundaries that align well with the GT.
To quantify the similarity, we compare Mutual Information (MI) values.
In scene 2 in \cref{fig:latent_pca}, StreamMapNet has an MI of 0.721, while AugMapNet has 0.761 (+\SI{5.5}{\percent}), which aligns with higher visual similarity.
For the whole nuScenes dataset, the latent BEV grid of StreamMapNet has an MI of $0.61\pm0.09$, while AugMapNet has $0.64\pm0.09$ (+\SI{4.9}{\percent}).
To show that higher similarity is related to improved vector map prediction performance, we plot the mAP over MI for all nuScenes val data points in \cref{fig:mutual_information} (note: data points with $0.67$ mAP are mostly scenes where the model predicts nonexistent crosswalks).
We measure a positive linear correlation between MI and vector map prediction performance, with $R^2=0.12$ for StreamMapNet and $R^2=0.22$ for AugMapNet.
To determine whether AugMapNet's increase of MI is statistically significant, we conduct a one-tailed t-test at a significance level $\alpha = 0.01$.
The test yields $p \ll 0.001$, thus, we reject the null hypothesis and conclude that MI is significantly higher in AugMapNet.
This result, in combination with the observed positive correlation between MI and vector map prediction performance, gives an interpretable insight and intuitive understanding of the performance improvement of AugMapNet.

Thirdly, a more structured latent BEV grid is expected to have an effect on its variance.
We calculate the spatial variance of each channel in the latent BEV grids and take an average across all channels.
For example, in scene 2 in \cref{fig:latent_pca}, the variance of StreamMapNet latent space is 0.103 \vs 0.068 for AugMapNet.
Results on the full nuScenes val dataset indicate that the latent space of AugMapNet indeed has a significantly lower variance ($0.055\pm0.006$) compared to StreamMapNet ($0.087\pm0.008$). 
We assume the variance reduction is caused by the BEV augmentation in AugMapNet. 
The raster map predicted in the BEV augmentation branch only carries low-dimensional class information due to the gradient stop from $e_{\mathrm{raster}}$ to $\mathcal{\hat{M}}_{\mathrm{raster}}$. 
Mapping this low-dimensional signal into the high-dimensional BEV grid latent space is expected to have a governing effect and reduce spatial variance.
An analysis of the relation between spatial variance of the latent BEV grid and vector map prediction performance is shown in \cref{fig:variance_analysis}.
In StreamMapNet, the variance does not correlate with prediction performance ($R^2=0.002$).
In contrast, AugMapNet shows a positive correlation ($R^2=0.143$), suggesting that its spatial variance is explained by relevant, task-specific features.

\subsection{Ablation of Latent BEV Grid Augmentation}\label{sec:ablation_bev_augmentation}
We evaluate our proposed latent BEV grid augmentation method and compare different variants of gradient stopping.
To allow a direct comparison with StreamMapNet, we do not include the additional processing CNNs (\ie, CNN1 and CNN2 in \cref{fig:augmapnet_architecture}).
The results on the nuScenes dataset are shown in \cref{tab:ablation_bev_augment}.
The baseline is StreamMapNet with \SI{33.8}{\percent}~mAP. 
First, we run an Oracle experiment to determine the upper bound of our BEV augmentation mechanism.
Here, the GT raster map, $\mathcal{M}_{\mathrm{raster}}$, not the predicted raster map, $\mathcal{\hat{M}}_{\mathrm{raster}}$, is passed to $e_{\mathrm{raster}}$; hence, the IoU scores are \SI{100.0}{\percent}.
It achieves \SI{91.4}{\percent} mAP, confirming our method's potential to improve vector map decoding.
{\begin{table}
    \centering
    \vspace{-2pt}
    \resizebox{\columnwidth}{!}{%
    \begin{tabular}{l|cc|cccc|c}
        Variant & $\nabla d$ & $\nabla e$ & AP$_{\mathrm{ped}}$ &  AP$_{\mathrm{div}}$ &  AP$_{\mathrm{bound}}$ & mAP & mIoU  \\ \midrule
        Baseline & - & - &  32.4 &  28.8 &  42.5 & 33.8 & - \\
        Oracle & - & - & 88.4 & 96.6 & 89.0 & 91.4 & 100.0 \\
        Full backprop & \cmark & \cmark &  32.9 &  29.3 &  42.3 & 34.8 & 41.4 \\
        $d_{\mathrm{raster}}$ grad stop & \xmark & \cmark & 30.8 & 28.1 & 41.5 & 33.5 & 27.3 \\
        $e_{\mathrm{raster}}$ grad stop & \cmark & \xmark &  33.3 &  29.1 &  \textbf{42.7} & 35.0 & \textbf{43.6} \\
        Isolated & \xmark & \xmark & \textbf{33.9} & \textbf{29.9} & \textbf{42.7} & \textbf{35.5} & 28.5 \\
    \end{tabular}%
    }
    \caption{Ablation of BEV augmentation on nuScenes. Oracle uses GT, $\mathcal{M}_{\mathrm{raster}}$, for augmentation. $\nabla$ indicates whether gradients are backpropagated from either $d_{\mathrm{raster}}$ or $e_{\mathrm{raster}}$ (\ie, dashed lines in \cref{fig:augmapnet_architecture}). AP and mIoU evaluate $\mathcal{\hat{M}}_{\mathrm{vector}}$ and $\mathcal{\hat{M}}_{\mathrm{raster}}$, respectively.}
	\label{tab:ablation_bev_augment}
\end{table}
}

The BEV augmentation mechanism introduces $d_{\mathrm{raster}}$ to decode $\mathcal{\hat{M}}_{\mathrm{raster}}$ from $B_{\mathrm{enc}}$ and $e_{\mathrm{raster}}$ to encode it to $B_{\mathrm{raster}}$.
With no gradient stopping (\enquote{Full backprop}), it gives a small improvement with \SI{34.8}{\percent}~mAP.
This variant also does raster map decoding well with \SI{41.4}{\percent}~mIoU.

Stopping the backpropagation between $d_{\mathrm{raster}}$ and $e_{\mathrm{BEV}}$ (\enquote{$d_{\mathrm{raster}}$ grad stop}) slightly drops vector map decoding performance to \SI{33.5}{\percent}~mAP.
This is expected, since $d_{\mathrm{raster}}$ is optimized by both $\mathcal{L}_\mathrm{vector}$ and $\mathcal{L}_\mathrm{raster}$, but has no influence on its input $B_{\mathrm{enc}}$.
As a result, a high-frequency latent space is likely formed by backpropagation.
A comparison with the \enquote{Isolated} case (see below) shows that this hurts performance.
Surprisingly, the raster map decoder still achieves \SI{27.3}{\percent} mIoU from a latent BEV grid that was only optimized for the vector map decoding task.
This finding suggests that decoding raster and vector representations requires similar latent features and that a hybrid approach can conceptualize abstract features useful for decoding both representations of the same underlying map information.

Alternatively, stopping the backpropagation from $e_{\mathrm{raster}}$ to $d_{\mathrm{raster}}$ (\enquote{$e_{\mathrm{raster}}$ grad stop}) improves performance to \SI{35.0}{\percent} mAP and \SI{43.6}{\percent} mIoU.
Finally, stopping backpropagation at both locations (\enquote{Isolated}) achieves the best result with \SI{35.5}{\percent} mAP.
\enquote{Isolated} is used in AugMapNet, since it cuts off the rasterized map branch from the rest of the model. 
Specifically, the latent BEV grid before augmentation, $B_{\mathrm{align}}$, is unaffected by the segmentation loss and can purely optimize for vector map decoding.
$\mathcal{\hat{M}}_{\mathrm{raster}}$ is learned separately and augments the latent BEV grid to improve the features for vector map decoding.
Supplementary~\ref{sec:gradstop_theory} provides a theoretical foundation for the empirical results of gradient stopping discussed in this section.

\subsection{Ablation of Latent BEV Grid Processing}\label{sec:ablation_bev_processing}
{\begin{table}
    \centering
    \vspace{-3pt}
    \resizebox{0.8\columnwidth}{!}{%
    \begin{tabular}{c|c|cccc}
        Index  & \# Layers  & AP$_{\mathrm{ped}}$ & AP$_{\mathrm{div}}$ & AP$_{\mathrm{bound}}$ & mAP \\ \midrule
        $a$)  & 0 & 31.2 & 27.3 & 42.9 & 33.8 \\ 
        $b$) & 1 & 36.4 & 30.0 & 43.7 & 36.7 \\
        $c$) & 2 & 38.9 & 29.6 & 43.5 & \textbf{37.3} \\
        $d$) & 3 & 36.7 & 30.0 & 43.1 & 36.6 \\
        $e$) & 4 & 34.3 & 30.0 & 44.8 & 36.4 \\
    \end{tabular}
    }
    \vspace{-3pt}
    \caption{Ablation of number of layers for BEV processing CNNs. $a$) is StreamMapNet \cite{streammapnet_2024} and $c$) is AugMapNet's setting.}
	\label{tab:ablation_temp_processing}
    \vspace{-5pt}
\end{table}
}
This study, shown in \cref{tab:ablation_temp_processing}, compares different variants for additional processing between the encoder and decoder.
Here we do not use our BEV augmentation to allow direct comparison with the StreamMapNet baseline, thus there is only one CNN block (\ie, CNN1 in \cref{fig:augmapnet_architecture}).
Comparing StreamMapNet ($a$) with index ($b$) shows that we get a significant improvement of \SI{8.6}{\percent} mAP just from adding a single convolution layer.
Another layer ($c$) gives a slight improvement, but further layers show a performance plateau ($d$, $e$).
Thus we choose 2 CNN layers ($c$) for AugMapNet.
Ablation on the kernel size is provided in Supplementary \ref{sec:ablation_kernel_size}.

\subsection{Ablation of Key Components of AugMapNet}\label{sec:ablation_key_component}
We assess the performance gains from AugMapNet's two key components: (1) latent BEV grid processing and (2) latent BEV grid augmentation.
As seen in \cref{tab:key_contribution}, BEV processing improves the StreamMapNet baseline by \SI{8.6}{\percent}.
In \cref{tab:ablation_temp_processing}, we saw that gains saturated at two CNN layers, with more layers adversely affecting results.
Despite this, our augmentation mechanism, which has 4 layers split across 2 CNN blocks, provides an additional \SI{4.4}{\percent} gain, bringing the full AugMapNet model to \SI{38.3}{\percent} mAP.
This additional improvement is even stronger for challenging long-range predictions: \SI{10.3}{\percent} on the $\SI{100}{\meter} \times \SI{50}{\meter}$ range and \SI{10.1}{\percent} on the $\SI{150}{\meter} \times \SI{75}{\meter}$ range. 
These results show the distinct roles of each component: latent BEV grid processing enhances the latent features, while augmentation refines the latent grid with dense spatial information, which is critical for long-range predictions. 
{\begin{table}[tbp]
    \centering
    \vspace{-2pt}
    \resizebox{0.9\columnwidth}{!}{%
    \begin{tabular}{l | l | l | l}
        Component & $\SI{60}{\meter} \times \SI{30}{\meter}$ & $\SI{100}{\meter} \times \SI{50}{\meter}$ & $\SI{150}{\meter} \times \SI{75}{\meter}$ \\
        \midrule
        Baseline & 33.8 & 23.2 & 6.9 \\
        + CNNs & 36.7 (+8.6\%) & 26.2 (+12.9\%) & 8.9 (+28.9\%) \\
        + Aug & \textbf{38.3} (+4.4\%) & \textbf{28.9} (+10.3\%) & \textbf{9.8} (+10.1\%) \\
    \end{tabular}
    }
    \caption{Ablation of key components of AugMapNet on nuScenes.}
    \label{tab:key_contribution}
    \vspace{-5pt}
\end{table}}

\section{Conclusion} \label{sec:conclusion}
We presented AugMapNet, a novel method for online vectorized HD map construction from multiple camera views. 
It enhances the latent BEV grid by further processing the features and then injecting spatial features from a predicted raster map, thus providing better dense spatial supervision for the vector decoder than other hybrid architectures.
Experiments on nuScenes and Argoverse2 show an improvement in mAP of up to \SI{13.3}{\percent} over the StreamMapNet baseline and even higher improvements on larger perception ranges with minimal parameter overhead.
AugMapNet shows particular strength in detecting complex map characteristics, such as pedestrian crossings, with a substantial \SI{26.3}{\percent} relative performance gain.

Additionally, our analysis of the latent BEV grid reveals that AugMapNet forms a more structured and task-relevant feature space.
Mutual Information analysis shows that AugMapNet’s augmented BEV grid has much higher similarity with the ground truth raster map, which is correlated with improved vector map predictions.

In conclusion, AugMapNet highlights the importance of dense spatial supervision in BEV-based vector map construction and provides an effective method to improve the latent structure generated by learned BEV encoders.

{
    \small
    \bibliographystyle{ieeenat_fullname}
    \bibliography{main}
}

\clearpage
\maketitlesupplementary
\appendix

\section*{Overview of Supplementary Material}\label{sec:supplementary}

The supplementary material includes the following parts:
\begin{enumerate}[label=\Alph*.]
\item Results on more Perception Ranges
\item Formulation of Dense Spatial Supervision
\item Mathematical Definition of Loss Functions
\item Theoretical Foundation for Gradient Stop
\item Non-functional Changes with AugMapNet Method
\item Results on Original nuScenes Split
\item Results on Argoverse2 for AugMapNet-SQD and Larger Range
\item Ablation of Kernel Size for Latent BEV Grid Processing CNNs
\item Additional Qualitative Results
\item Visualization of Principal Components
\end{enumerate}

\section{Results on more Perception Ranges}\label{sec:ablation_range}
\cref{tab:ablation_perception_ranges_full} shows the results on various perception ranges.
Beyond the reported \SI{+13.3}{\percent} improvement on perception range $\SI{60}{\meter} \times \SI{30}{\meter}$, we get \SI{+19.6}{\percent} on $\SI{80}{\meter} \times \SI{40}{\meter}$, \SI{+24.6}{\percent} on $\SI{100}{\meter} \times \SI{50}{\meter}$, \SI{+23.4}{\percent} on $\SI{120}{\meter} \times \SI{60}{\meter}$, and \SI{+41.4}{\percent} on $\SI{150}{\meter} \times \SI{75}{\meter}$.
We find that the relative improvement of AugMapNet increases with larger perception ranges, which suggests a stronger benefit of our augmentation method on wider ranges.

\section{Formulation of Dense Spatial Supervision}\label{sec:spatial_formulation_supplementary}
The loss is calculated over all elements of the output representation $\hat{\mathcal{M}}$.
Supervision is done by optimizing a learnable set of weights $W$ based on the gradients of the loss:
\begin{equation}
\frac{\partial L}{\partial W} = \frac{\partial L}{\partial \hat{\mathcal{M}}}  \frac{\partial \hat{\mathcal{M}}}{\partial W}.
\end{equation}
The total amount of spatial supervision therefore is related to the number of elements in $\hat{\mathcal{M}}$.

In the case of a raster map decoder, the predicted output is a rasterized representation: $\hat{\mathcal{M}} = \hat{\mathcal{M}}_{\mathrm{raster}}$.
The total gradient signal is calculated from the loss over all pixels.
For a raster with spatial resolution $H=100$ and $W=50$, this gives a spatial supervision for $| \hat{\mathcal{M}}_{\mathrm{raster}} | = H \times W = 100 \times 50 = 5000$ elements.

In the case of a vector map decoder, the predicted output is a vectorized representation:
$\hat{\mathcal{M}} = \hat{\mathcal{M}}_{\mathrm{vector}}$. 
The total gradient signal is calculated from the loss over all points of all polylines.
Our model outputs $N=100$ polylines, with $N_p = 20$ points each, which gives spatial supervision for at most $| \hat{\mathcal{M}}_{\mathrm{vector}} | = N \times N_p = 100 \times 20 = 2000$ elements.
Note that, in DETR-style decoders, elements that are correctly predicted as empty class do not induce spatial supervision.
Hence, this number is the upper bound and the actual supervision is typically even sparser.

We believe that this difference in the number of elements is why integrating dense spatial supervision into vectorized map decoding is so effective.

{\begin{table}
    \centering
    \resizebox{\columnwidth}{!}{%
    \begin{tabular}{ll|cccc|c}
        Range & Variant & AP$_{\mathrm{ped}}$ &  AP$_{\mathrm{div}}$ &  AP$_{\mathrm{bound}}$ & mAP & Impr. \\ \midrule
        \multirow{2}{*}{$\SI{60}{\meter} \times \SI{30}{\meter}^\ddagger$} & StreamMapNet \cite{streammapnet_2024} &  31.2 & 27.3 & 42.9 & 33.8 &  \\
        & AugMapNet (ours)  & 39.4 & 30.3 & 45.3 & 38.3 & \SI{+13.3}{\percent} \\ \midrule
        \multirow{2}{*}{$\SI{80}{\meter} \times \SI{40}{\meter}^\ddagger$} & StreamMapNet \cite{streammapnet_2024} & 19.4 & 18.9  & 24.1  & 20.8 &  \\
        & AugMapNet (ours)  & 25.9 & 21.1  & 27.6  & 24.9 & \SI{+19.6}{\percent} \\ \midrule
        \multirow{2}{*}{$\SI{100}{\meter} \times \SI{50}{\meter}^\ast$} & StreamMapNet \cite{streammapnet_2024} & 25.5 & 19.3  & 24.7  & 23.2 &  \\
        & AugMapNet (ours) & 35.5 & 22.8  & 28.4  & 28.9 & \SI{+24.6}{\percent}\\ \midrule
        \multirow{2}{*}{$\SI{120}{\meter} \times \SI{60}{\meter}^\ast$} & StreamMapNet \cite{streammapnet_2024} & 19.29 & 10.95  &  12.79  & 14.34 &  \\
        & AugMapNet (ours) & 24.13 & 12.72  & 16.26   &  17.70 & \SI{+23.4}{\percent} \\ \midrule
        \multirow{2}{*}{$\SI{150}{\meter} \times \SI{75}{\meter}^\ast$} & StreamMapNet \cite{streammapnet_2024} & 10.5 & 6.4 & 3.9  & 6.9 &  \\ 
        & AugMapNet (ours) & 15.2 & 7.5  & 6.7  & 9.8 & \SI{+41.4}{\percent} \\
    \end{tabular}%
    }
    \caption{Results on nuScenes dataset \cite{nuscenes} on more perception ranges. AP thresholds $^\ddagger$: \{0.5, 1.0, 1.5\}, $^\ast$: \{1.0, 1.5, 2.0\}.}
    \label{tab:ablation_perception_ranges_full}
\end{table}
}

\section{Mathematical Definition of Loss Functions}\label{sec:loss_supplementary}

The formulation of the vector map decoder loss, $\mathcal{L}_\mathrm{vector}$, is taken from StreamMapNet \cite{streammapnet_2024} and consists of multiple components. 
As the first step of the polyline matching loss, $\mathcal{L}_\mathrm{line}$, bipartite matching is performed between predicted and GT polylines.
After matching, the smooth L1 loss is calculated for each of the $N_p$ points $P_j$ of the matched polyline $P$.
The best one in the permutation group, $\Gamma$, as introduced in MapTR \cite{maptrv2_2023}, is used:
\begin{equation}
    \mathcal{L}_\mathrm{line}(\hat{P}, P) = 
    \min_{\gamma \in \Gamma}
    \frac{1}{N_p}\sum_{j=1}^{N_p} 
    \mathcal{L}_\mathrm{SmoothL1}(\hat{P}_j, P_{\gamma(j)}).
\end{equation}

The classification loss, $\mathcal{L}_\mathrm{class}$, calculates the loss between the ground truth class vector, $c$, and predicted class vector, $\hat{c}$, for each polyline.
The loss function is the Focal loss:
\begin{equation}
    \mathcal{L}_\mathrm{class}(\hat{c}, c) = \mathcal{L}_\mathrm{Focal}(\hat{c}, c).
\end{equation}

An auxiliary transformation loss, $\mathcal{L}_\mathrm{trans}$, is used to match the ego-motion transformation in latent space.
Given a standard $4 \times 4$ transformation matrix, $T$, between the coordinate frames of $t-1$ and $t$, the polyline of the vector map at time $t$ is expressed as $P =  T\cdot \mathrm{homogeneous}(P')_{:,0:2} \, $, where $P'$ is the polyline of the vector map at time $t-1$.
For a query in the vector map decoder at time $t$, $Q$, an auxiliary prediction is made with $\hat{P}^{\, \mathrm{aux}} = \mathrm{Reg}(Q)$.
The auxiliary transformation loss is then defined as:
\begin{equation}
    \mathcal{L}_\mathrm{trans} (\hat{P}^{\, \mathrm{aux}}, P) = \sum_{j=1}^{N_p} 
    \mathcal{L}_\mathrm{SmoothL1}(\hat{P}^{\, \mathrm{aux}}_j, P_j).
\end{equation}

The final loss is a weighted sum of the above loss terms with the factors $\lambda_{1}=50.0$, $\lambda_{2}=5.0$, and $\lambda_{3}=0.1$ over all predicted polylines in $\hat{\mathcal{M}}_{\mathrm{vector}}$:
\begin{equation}
    \mathcal{L}_\mathrm{vector}=
    \sum_{P \in \hat{\mathcal{M}}_{\mathrm{vector}}} 
    \left( \lambda_1\mathcal{L}_\mathrm{line}
    +\lambda_2\ \mathcal{L}_\mathrm{class}
    +\lambda_3 \mathcal{L}_\mathrm{trans} \right).
    \label{eq:loss_vector}
\end{equation}

The raster map decoder loss $\mathcal{L}_\mathrm{raster}$ is the Dice loss \cite{milletari2016dice}:
\begin{equation}
    \mathcal{L}_\mathrm{raster}(\hat{\mathcal{M}}_{\mathrm{seg}}, \mathcal{M}_{\mathrm{seg}}) = \mathcal{L}_\mathrm{Dice}(\hat{\mathcal{M}}_{\mathrm{seg}}, \mathcal{M}_{\mathrm{seg}}).
\end{equation}

\section{Theoretical Foundation for Gradient Stop}\label{sec:gradstop_theory}
We provide more analysis on gradient stopping to provide a theoretical foundation for the benefits that we observed empirically.
We start our reasoning from the results of the \enquote{Oracle} experiment (see details of experiment in \cref{sec:ablation_bev_augmentation}).

The result of \SI{91.4}{\percent}~mAP shows that given a perfect raster map, the injection of dense spatial features in our AugMapNet method can yield a close-to-perfect result on the vectorized map construction.
We denote using this perfect raster map, that reaches the \emph{empirical upper bound} in the \enquote{Oracle} experiment, as $\hat{\mathcal{M}}_{\mathrm{raster}}=\mathcal{M}_{\mathrm{raster}}^{*}$.
In practice, the predicted raster map includes residual error $\eta$, such that $\hat{\mathcal{M}}_{\mathrm{raster}} = \mathcal{M}_{\mathrm{raster}}^{*} + \eta$.
Without gradient stopping $\nabla e_\mathrm{{raster}}$ is active, backpropagating gradients from the vector loss $\mathcal{L}_{\mathrm{vector}}$ to $d_\mathrm{{raster}}$.
Therefore, instead of purely optimizing for a perfect raster map, $\frac{\partial \mathcal{L}_{\mathrm{vector}}}{\partial W_{d_{\mathrm{raster}}}}$ biases $d_{\mathrm{raster}}$ toward vector‑specific features, increasing $ \|\eta\| $.

\section{Non-functional Changes with AugMapNet Method}\label{sec:non-functional}
This section provides more details on the non-functional changes when applying our AugMapNet method to StreamMapNet.
Our AugMapNet method has only a small effect on the model size in VRAM. We observe \SI{\approx 1.8}{\giga\byte} for either of them.
During training on batch size 1, we see a VRAM usage of \SI{\approx 7.7}{\giga\byte}, a slight increase of \SI{+1.6}{\percent}.
Train time also increases slightly from \SI{14.5}{\hour} to \SI{15.4}{\hour} (\SI{+6}{\percent}).
Inference takes \SI{78}{\milli\second} \vs \SI{70}{\milli\second}, an \SI{11}{\percent} increase.
AugMapNet still meets real-time requirements with 11.9~FPS.

\section{Results on Original nuScenes Split}\label{sec:results_original_nuscenes}

{\begin{table}[t]
    \resizebox{\columnwidth}{!}{%
    \begin{tabular}{l|cccc|c}
        Method & AP$_{\mathrm{ped}}$&  AP$_{\mathrm{div}}$  &  AP$_{\mathrm{bound}}$& mAP  & Impr. \\ 
        \midrule
        VectorMapNet \cite{vectormapnet_2023} & 36.1 & 47.3 & 39.3 & 40.9 \\
        MapTR \cite{maptr_2023} & 46.3 & 51.5 & 53.1 & 50.3 \\
        MapTRv2 \cite{maptrv2_2023} & 59.8 & 62.4 & 62.4 & 61.5 \\
        MapVR \cite{zhang2024mapvr} & 47.7 & 54.5 & 51.4 & 51.2 \\
        MGMap \cite{liu2024mgmap} & 57.4 & 63.5 & 63.3 & 61.4 \\
        \midrule
        StreamMapNet \cite{streammapnet_2024} & {60.2} & {65.1} & {61.1} & {62.1} \\
        AugMapNet (ours, \textcolor{teal}{$\nabla e_\mathrm{{raster}}$} off) & {60.8} & {65.7} & {61.7} & {62.7} & \SI{+1.0}{\percent} \\
        AugMapNet (ours, \textcolor{teal}{$\nabla e_\mathrm{{raster}}$} on) & {61.9} & {65.4} & {63.6} & {63.6} & \SI{+2.4}{\percent} \\
        StreamMapNet 100x50 \cite{streammapnet_2024} & {63.5} & {64.9} & {57.1} & {61.8} \\
        AugMapNet 100x50 (ours) & {66.1} & {64.1} & {60.0} & {63.4} & \SI{+2.5}{\percent} \\
        \midrule
        SQD-MapNet \cite{wang2024stream} & {62.2} & {67.0} & {65.4} & {64.9} \\
        AugMapNet-SQD (ours) & {64.1} & {65.0} & {67.8} & {65.6} & \SI{+1.1}{\percent} \\
        SQD-MapNet 100x50 \cite{wang2024stream} & {62.9} & {65.8} & {61.4} & {63.3} \\
        AugMapNet-SQD 100x50 (ours) & {67.3} & {68.6} & {63.7} & {66.5} & \SI{+3.5}{\percent} \\ 
    \end{tabular}%
    }
    \caption{Results on original nuScenes \cite{nuscenes} split with geospatial overlap. Results from baselines are taken from the respective papers. Values are much higher compared to geospatially disjoint split due to overfitting.}
    \label{tab:results_nuscenes_original}
\end{table}
}

The original nuScenes Split is still commonly used to train and evaluate approaches for online map construction.
For completeness, the results on the original split are shown in \cref{tab:results_nuscenes_original}.
Comparing the absolute values to the results on the geospatially disjoint split (\cf \cref{tab:table_main_result}) reveals severe overfitting due to geospatial overlap: StreamMapNet result increases from \SI{33.8}{\percent}~mAP to \SI{62.1}{\percent}~mAP, an \SI{84}{\percent} jump due to overfitting.
AugMapNet overfits less with only \SI{64}{\percent} higher mAP (\SI{62.7}{\percent}~mAP on old \vs \SI{38.3}{\percent}~mAP on geospatially disjoint splits). 
Thanks to less overfitting, AugMapNet improvements are smaller on the old split, though still substantial.
A primary reason is gradient stopping: turning off \textcolor{teal}{$\nabla e_\mathrm{{raster}}$} improves performance on the geospatially disjoint split
(\cf \cref{sec:ablation_bev_augmentation}) but not on the old split (\SI{62.7}{\percent}~mAP without \textcolor{teal}{$\nabla e_\mathrm{{raster}}$} \vs \SI{63.6}{\percent}~mAP without \textcolor{teal}{$\nabla e_\mathrm{{raster}}$}).
Consistent with the results on the geospatially disjoint split, AugMapNet shows an even stronger improvement on larger perception ranges with \SI{2.5}{\percent} when evaluated on $\SI{100}{\meter} \times \SI{50}{\meter}$ range on the original split (\cf \cref{tab:ablation_perception_ranges} and \cref{tab:results_nuscenes_original}).

We follow best practices in ML to only perform comparisons on truly unseen evaluation data (\ie, the geospatially disjoint split).
For completeness, other baselines are added to \cref{tab:results_nuscenes_original}.
We note that comparison of absolute values mainly depends on the choice of baseline.
Once code for the latest state-of-the-art models is available, AugMapNet can easily be integrated to further advance them. 

\section{Results on Argoverse2 for AugMapNet-SQD and Larger Range}\label{sec:results_augmapnet-sqd_argoverse2}

{\begin{table}[t]
    \centering
    \resizebox{\columnwidth}{!}{%
    \begin{tabular}{ll|cccc|c}
        Range & Method & AP$_{\mathrm{ped}}$&  AP$_{\mathrm{div}}$  &  AP$_{\mathrm{bound}}$& mAP  & Impr. \\ 
        \midrule
        \multirow{2}{*}{$\SI{60}{\meter}\!\times\!\SI{30}{\meter}^\ddagger$} & StreamMapNet \cite{streammapnet_2024} &  56.0 & 54.4 & 61.0 & 57.1 & \\
        & AugMapNet (ours) & 57.4 & 57.4 & 61.6 & 58.8 & \SI{+3.0}{\percent} \\
        \multirow{2}{*}{$\SI{100}{\meter}\!\times\!\SI{50}{\meter}^\ast$} & StreamMapNet \cite{streammapnet_2024} & 57.9 & 44.4 & 47.5 & 49.9 & \\
        & AugMapNet (ours) & 60.6 & 44.8 & 49.4 & 51.6 & \SI{+3.4}{\percent} \\
        \midrule
        \multirow{2}{*}{$\SI{60}{\meter}\!\times\!\SI{30}{\meter}^\ddagger$} & SQD-MapNet \cite{wang2024stream} & 58.3& 54.7& 62.2& 58.4& \\
        & AugSQD (ours) & 60.0& 58.0& 64.6& 60.9& \SI{+4.3}{\percent}\\
        \multirow{2}{*}{$\SI{100}{\meter}\!\times\!\SI{50}{\meter}^\ast$} & SQD-MapNet \cite{wang2024stream} & 59.2& 45.9& 49.7& 51.6& \\
        & AugSQD (ours) & 63.3& 49.7& 51.4& 54.8& \SI{+6.2}{\percent}\\
    \end{tabular}
    }
    \caption{Results on different perception ranges for AugMapNet and application of our method to SQD-MapNet \cite{wang2024stream} denoted as AugSQD. Results are on Argoverse2 split without geospatial overlap \cite{nuscenes}. AP thresholds $^\ddagger$: \{0.5, 1.0, 1.5\}, $^\ast$: \{1.0, 1.5, 2.0\}.}
    \label{tab:table_sqd_argo}
\end{table}
}

This section extends upon the results of AugMapNet on Argoverse2 (see \cref{tab:table_argo_result}).
To assess how well our method generalizes across datasets and sensor configurations on different ranges, we extend \cref{tab:table_argo_result} by adding $\SI{100}{\meter} \times \SI{50}{\meter}$ range.
Beyond the \SI{3.0}{\percent} improvement on $\SI{60}{\meter} \times \SI{30}{\meter}$ range, we observe \SI{3.4}{\percent} improvement on $\SI{100}{\meter} \times \SI{50}{\meter}$ range.

To assess how well our method generalizes not only across datasets and sensor configurations, but also across models without any changes, we train AugMapNet-SQD (\enquote{AugSQD}) on the Argoverse2 dataset \cite{argoverse} and show the results in \cref{tab:table_sqd_argo}.
AugMapNet-SQD reaches \SI{60.9}{\percent}~mAP on $\SI{60}{\meter} \times \SI{30}{\meter}$ range, a \SI{4.3}{\percent} improvement over SQD-MapNet. 
On $\SI{100}{\meter} \times \SI{50}{\meter}$ range, AugMapNet-SQD reaches \SI{54.8}{\percent}~mAP, an even higher improvement of \SI{6.2}{\percent}
The improvements are substantial and even higher than the \SI{3.0}{\percent} improvement of AugMapNet over StreamMapNet on Argoverse2 (\cf \cref{tab:table_argo_result}), confirming the broader applicability of our method. 
Furthermore, the values show an increase in relative improvement at larger perception ranges a fifth time (\cf \cref{tab:ablation_perception_ranges} and \cref{tab:results_nuscenes_original}).

\section{Ablation of Kernel Size for Latent BEV Grid Processing CNNs} \label{sec:ablation_kernel_size}
In this study, shown in \cref{tab:ablation_kernel_size}, we investigate the effect of kernel size by comparing sizes of 1, 3, and 5 ($b$, $c$, $d$, resp.) on nuScenes.
Similar to \cref{sec:ablation_bev_processing}, here we do not use our BEV augmentation.
Size 3 is used for further experiments since it gives the best result with \SI{36.7}{\percent} mAP.

{\begin{table}
    \centering
    \resizebox{0.9\columnwidth}{!}{%
    \begin{tabular}{ccc|cccc}
        Index & Kernel & \# Layers  & AP$_{\mathrm{ped}}$ & AP$_{\mathrm{div}}$ & AP$_{\mathrm{bound}}$ & mAP \\ \midrule
        $a$) & - & 0 & 31.2 & 27.3 & 42.9 & 33.8 \\
        $b$) & 1 & 1 & 36.6 & 30.3 & 41.4 & 36.1 \\
        $c$) & 3 & 1 & 36.4 & 30.0 & 43.7 & \textbf{36.7} \\
        $d$) & 5 & 1 & 33.9 & 30.3 & 43.8 & 36.0 \\
    \end{tabular}
    }
    \caption{Ablation of kernel size of BEV processing CNNs. $a$) is StreamMapNet \cite{streammapnet_2024}.}
	\label{tab:ablation_kernel_size}
\end{table}
}

\section{Additional Qualitative Results}\label{sec:qualitative_results_extended}
We provide further qualitative results to highlight the benefits of AugMapNet.

A separate video file is included with this submission to show the performance of AugMapNet on a full scene.
It overlays the predicted vector map polylines in each camera view to visualize the spatial accuracy.

\begin{figure*}
    \centering
    \includegraphics[width=0.97\textwidth]{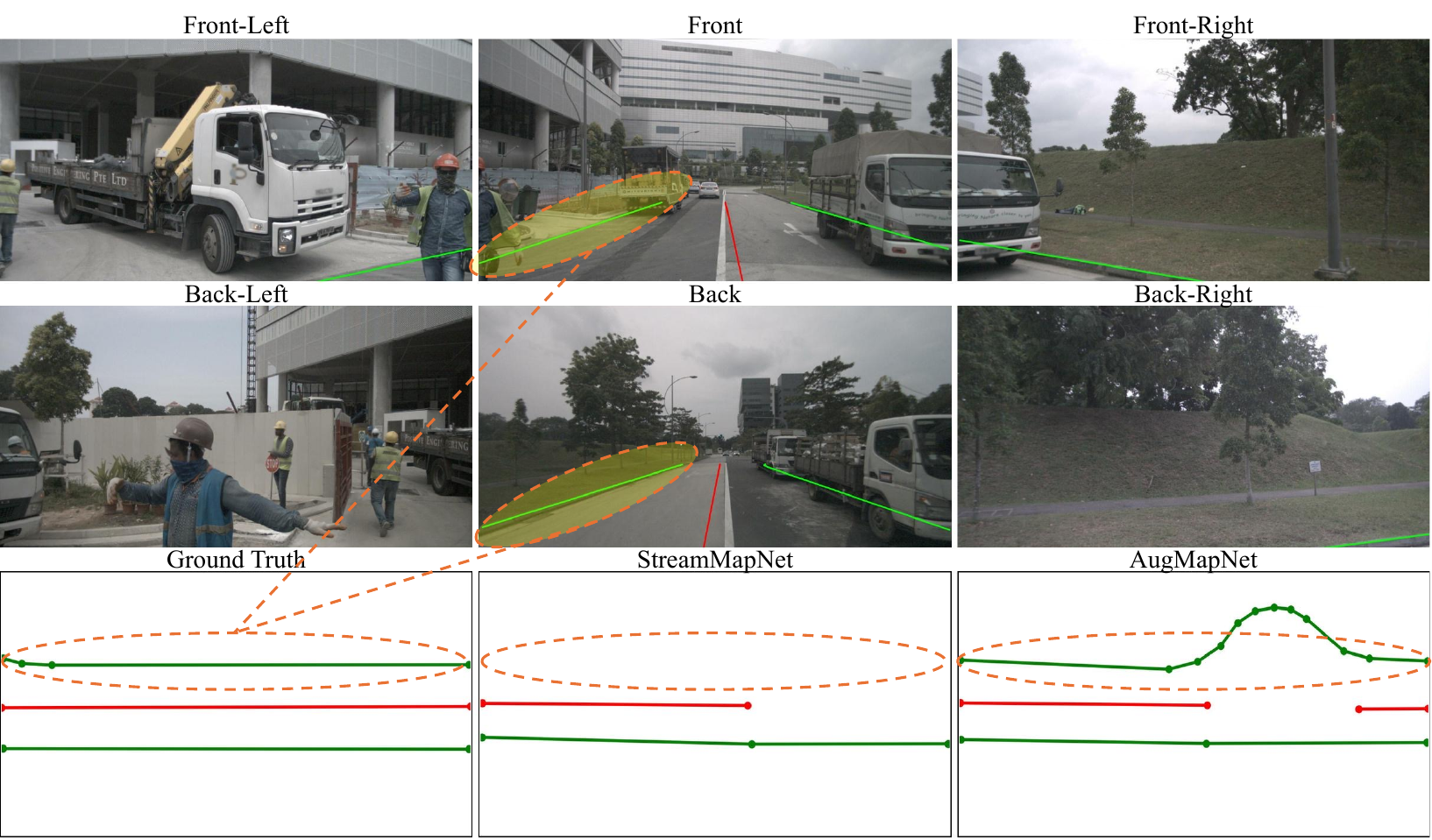}    
   \caption{Qualitative results for example scene 3 with an extensive driveway on the left side. Lane dividers are red, road boundaries are green, and pedestrian crossings are blue.}
   \label{fig:supp_qualitative_results_ex3}
\end{figure*}

\begin{figure*}
    \centering
    \includegraphics[width=0.97\textwidth]{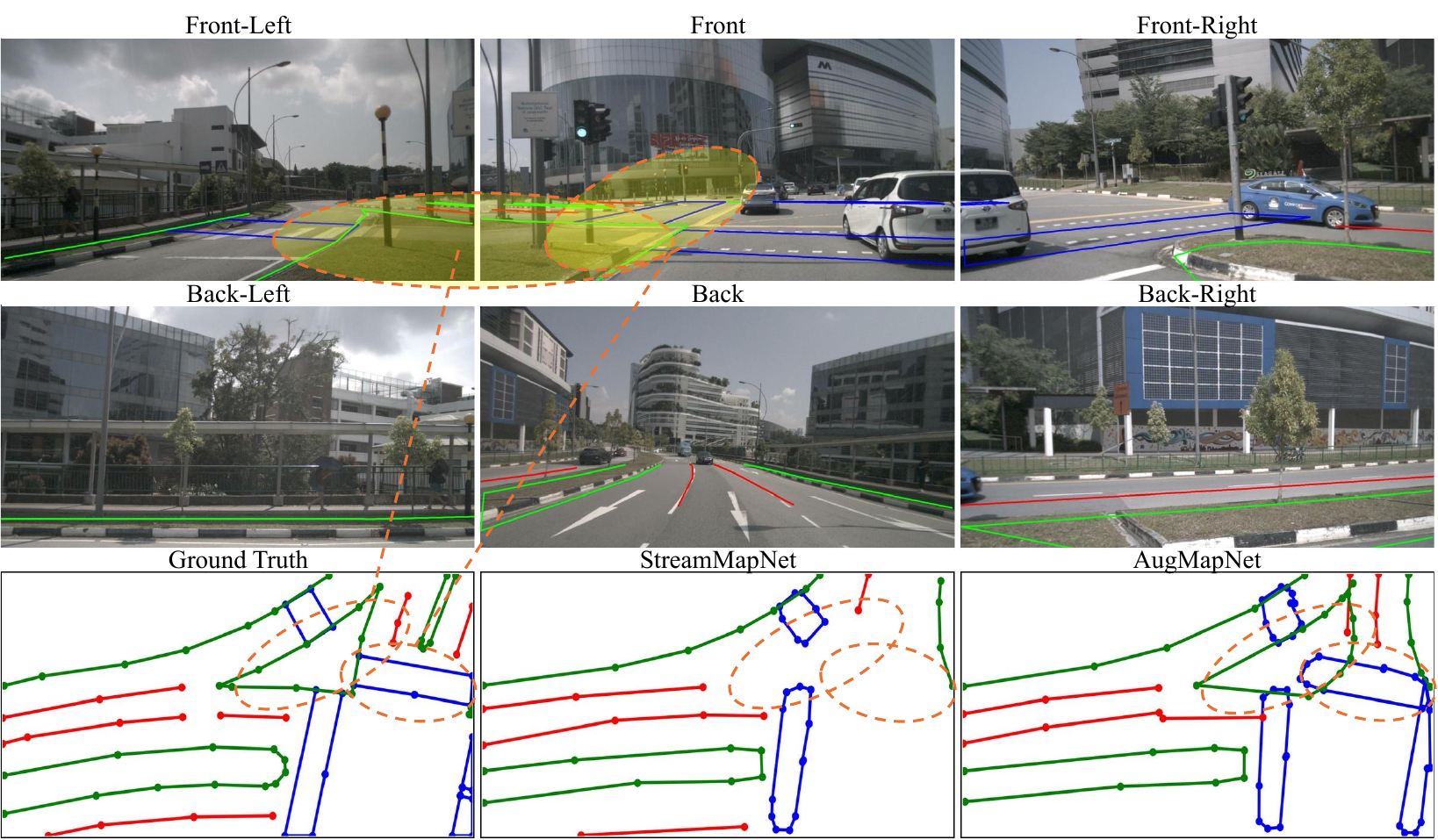}    
   \caption{Qualitative results for example scene 4 with traffic island and pedestrian crossing on the left. Lane dividers are red, road boundaries are green, and pedestrian crossings are blue.}
   \label{fig:supp_qualitative_results_ex4}
\end{figure*}

\begin{figure*}
    \centering
    \includegraphics[width=0.97\textwidth]{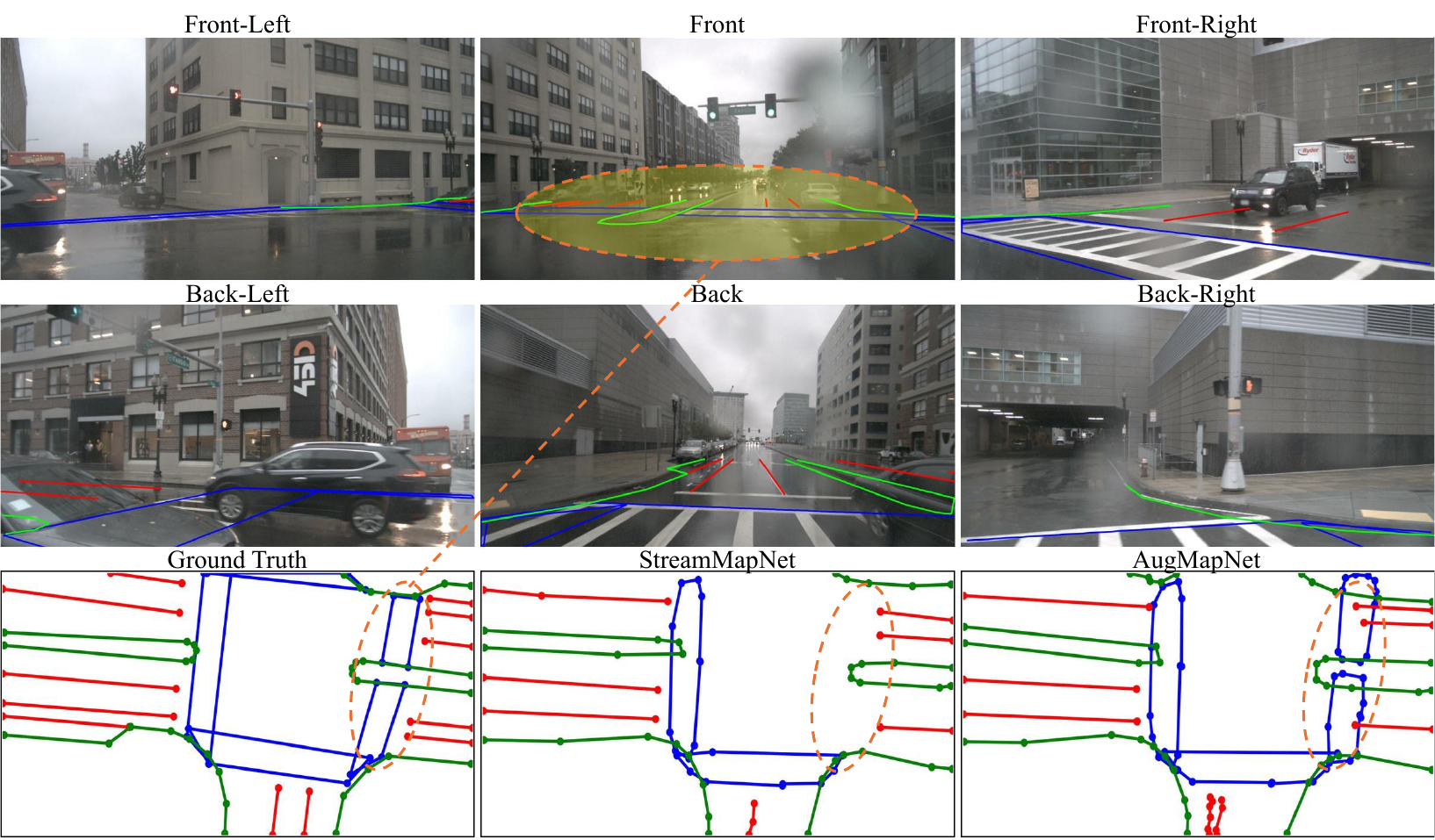}    
   \caption{Qualitative results for example scene 5 under rainy conditions. Lane dividers are red, road boundaries are green, and pedestrian crossings are blue.}
   \label{fig:supp_qualitative_results_ex5}
\end{figure*}

\begin{figure*}
    \centering
    \includegraphics[width=0.97\textwidth]{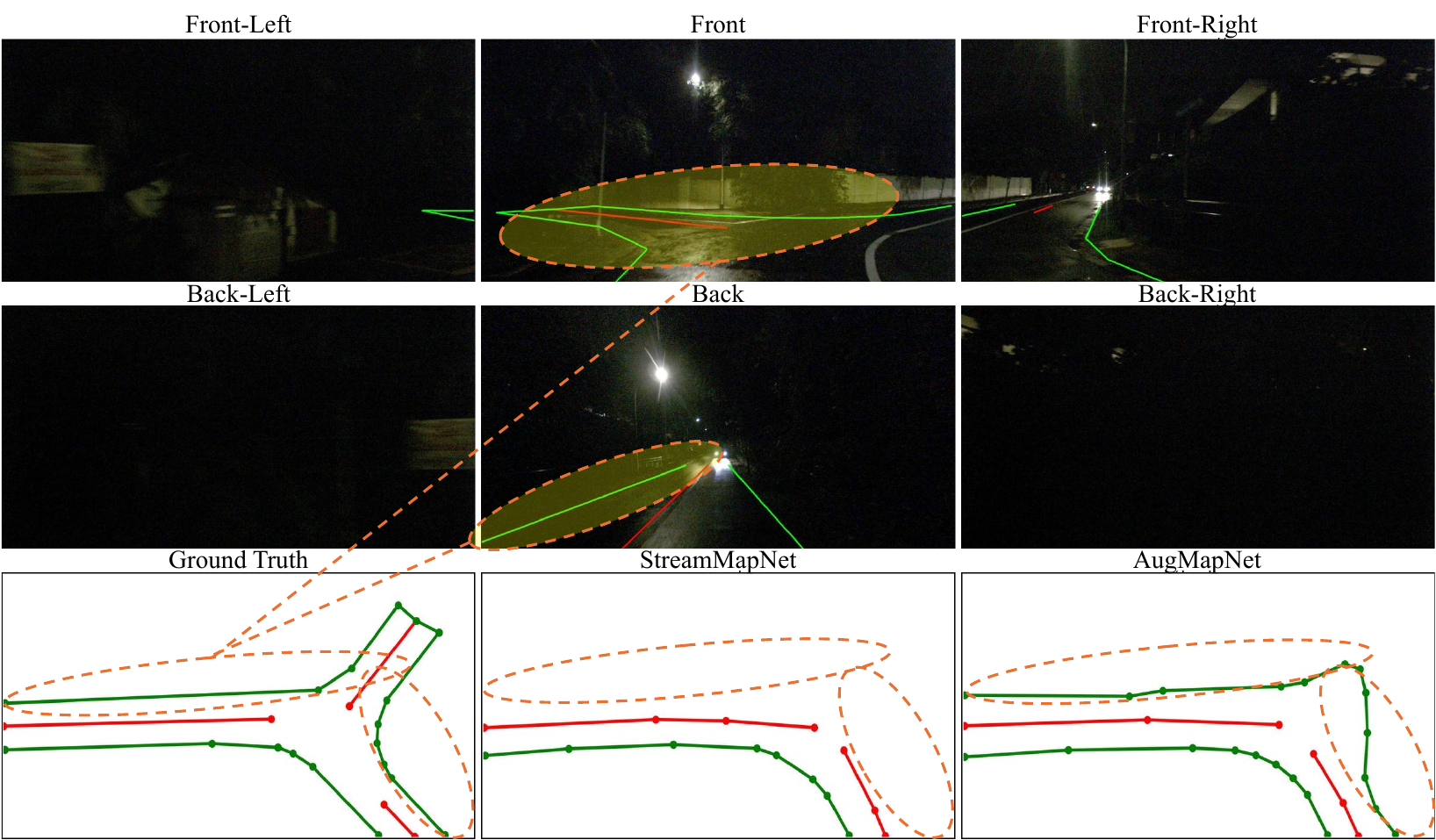}    
   \caption{Qualitative results for example scene 6 with limited illumination. Lane dividers are red, road boundaries are green, and pedestrian crossings are blue.}
   \label{fig:supp_qualitative_results_ex6}
\end{figure*}

\cref{fig:supp_qualitative_results_ex3} shows a scene where StreamMapNet misses the road divider on the left side of the ego vehicle.
AugMapNet predicts the road divider to include a roadway turnout, which is reasonable given the large driveway that is occupied by a truck.
\cref{fig:supp_qualitative_results_ex4} shows a scene where a traffic island and a crosswalk to the front-left of the ego vehicle are missed by StreamMapNet, but correctly predicted by AugMapNet.
\cref{fig:supp_qualitative_results_ex5} visualizes a rainy scene to show model performance under challenging weather conditions, including a rain droplet on the front-facing camera lens that limits visibility.
StreamMapNet misses a crosswalk in the area that AugMapNet predicts correctly.
\cref{fig:supp_qualitative_results_ex6} shows a scene at night with limited illumination.
AugMapNet predicts the left road boundary fairly accurately given the visibility, whereas StreamMapNet misses it completely.

\section{Visualization of Principal Components}\label{sec:pca_plots}

\begin{figure*}
    \centering
    \begin{subfigure}[t]{0.8\textwidth}
        \centering
        \includegraphics[width=\textwidth]{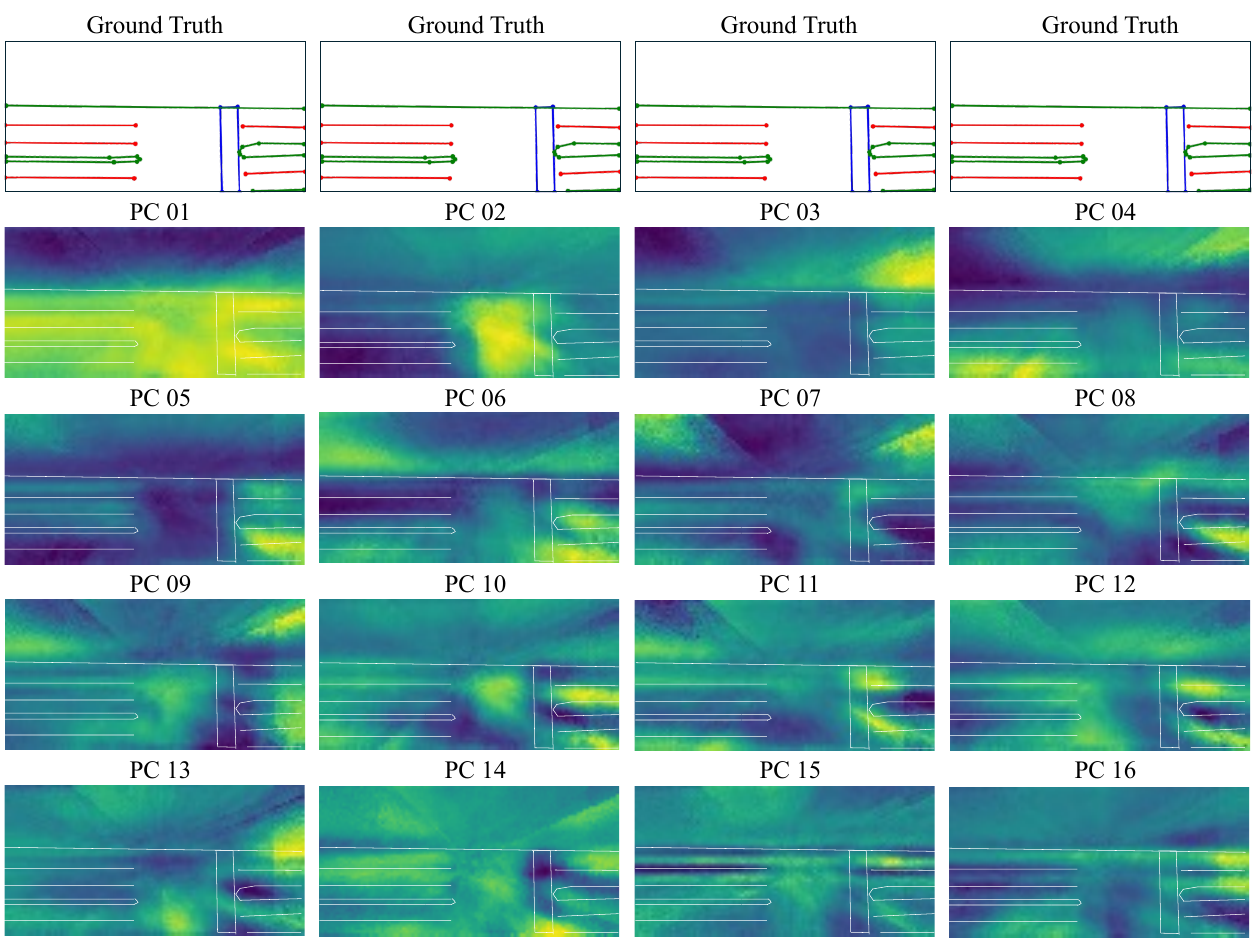}
        \caption{StreamMapNet}
        \label{fig:top16_pc_smn_ex1}
    \end{subfigure}
    \hfill
    \begin{subfigure}[t]{0.8\textwidth}
        \centering
        \includegraphics[width=\textwidth]{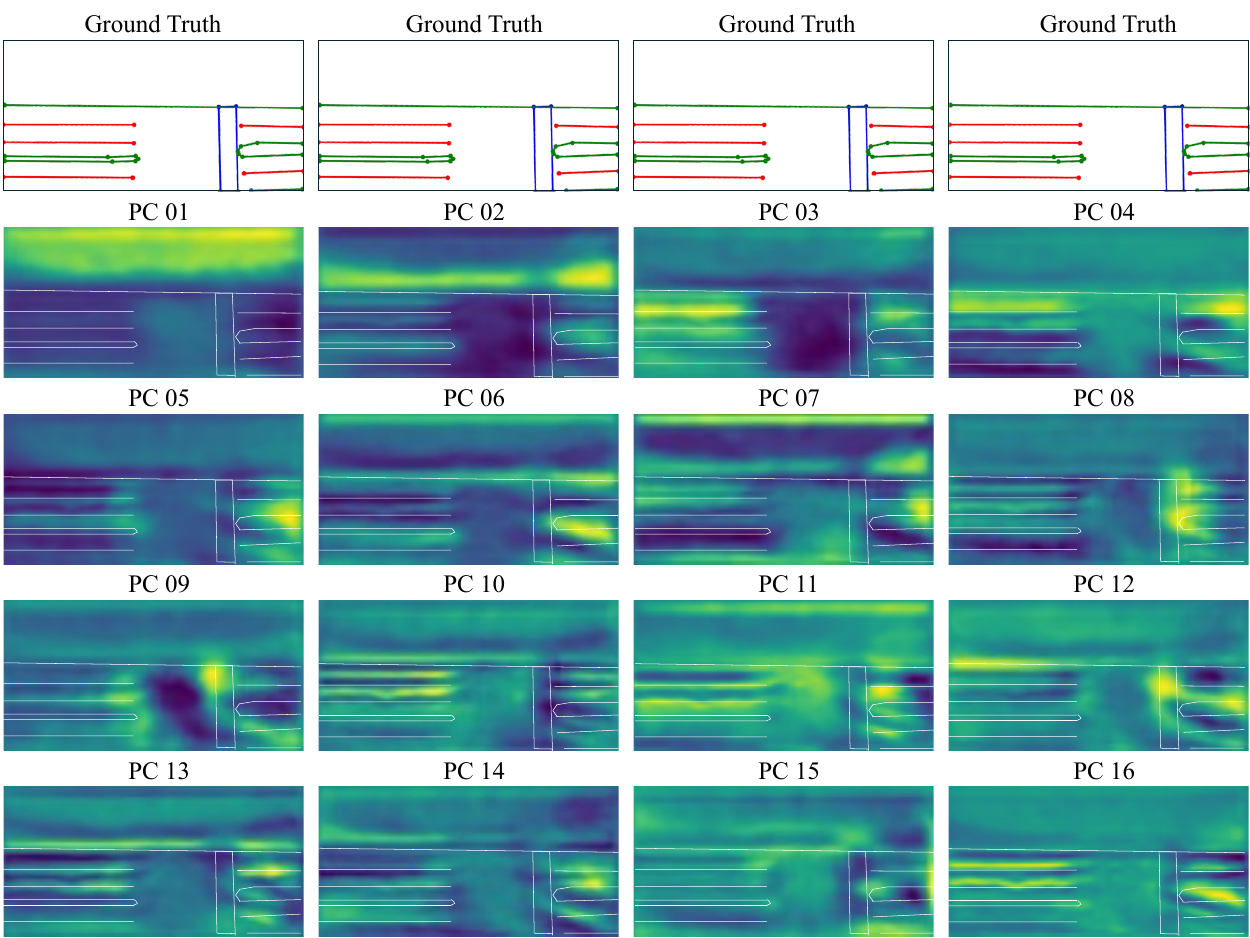}
        \caption{AugMapNet}
        \label{fig:top16_pc_amn_ex1}
    \end{subfigure}
   \caption{Example scene 1: Top 16 Principal Components of latent BEV grid of (a) StreamMapNet and (b) AugMapNet. Vector map ground truth is visualized at the top of each column as well as overlaid as white lines to ease the assessment of spatial accuracy. Notice that StreamMapNet tends to have radial artifacts.}
   \label{fig:top16_pc_ex1}
\end{figure*}

\begin{figure*}
    \centering
    \begin{subfigure}[t]{0.8\textwidth}
        \centering
        \includegraphics[width=\textwidth]{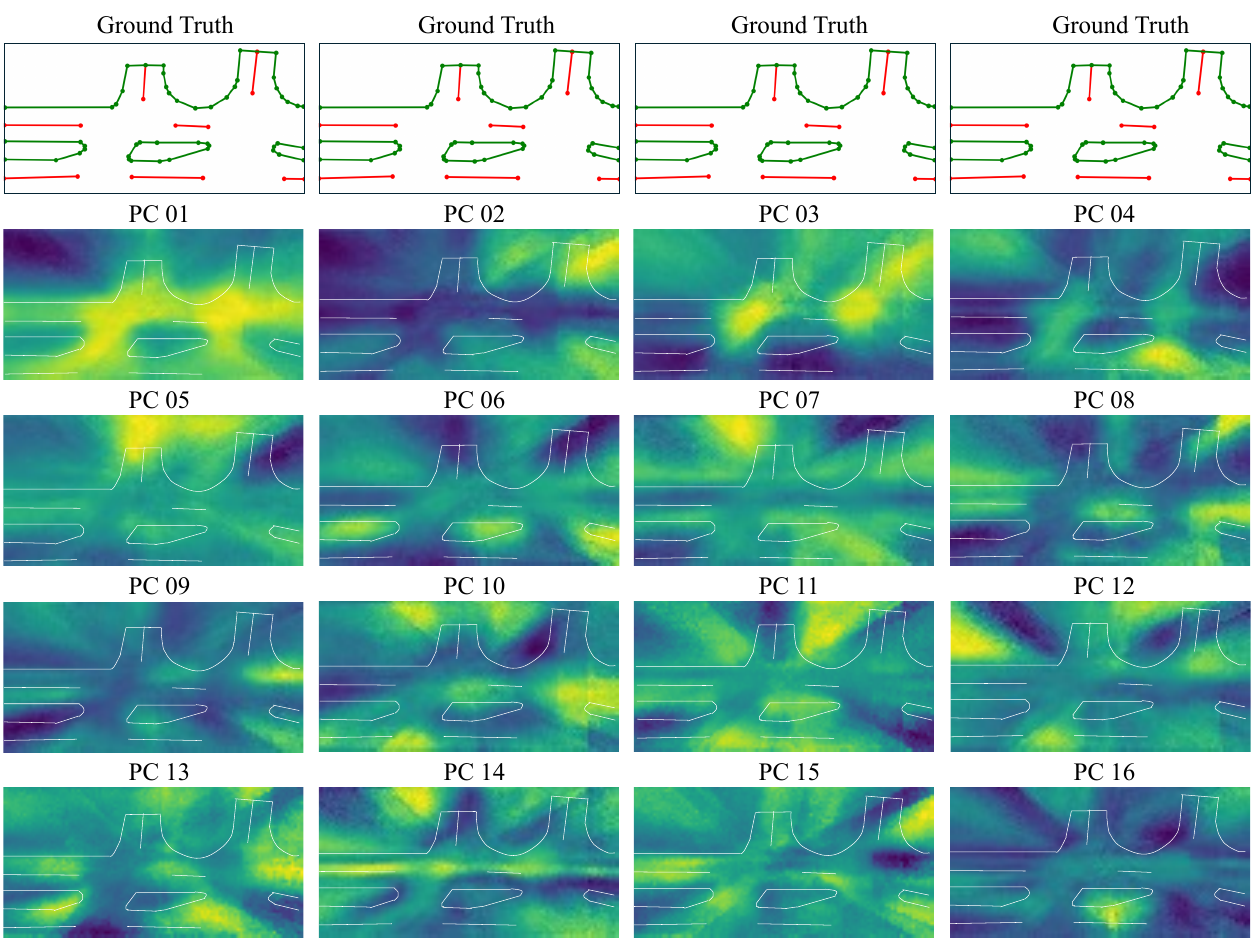}
        \caption{StreamMapNet}
        \label{fig:top16_pc_smn_ex2}
    \end{subfigure}
    \hfill
    \begin{subfigure}[t]{0.8\textwidth}
        \centering
        \includegraphics[width=\textwidth]{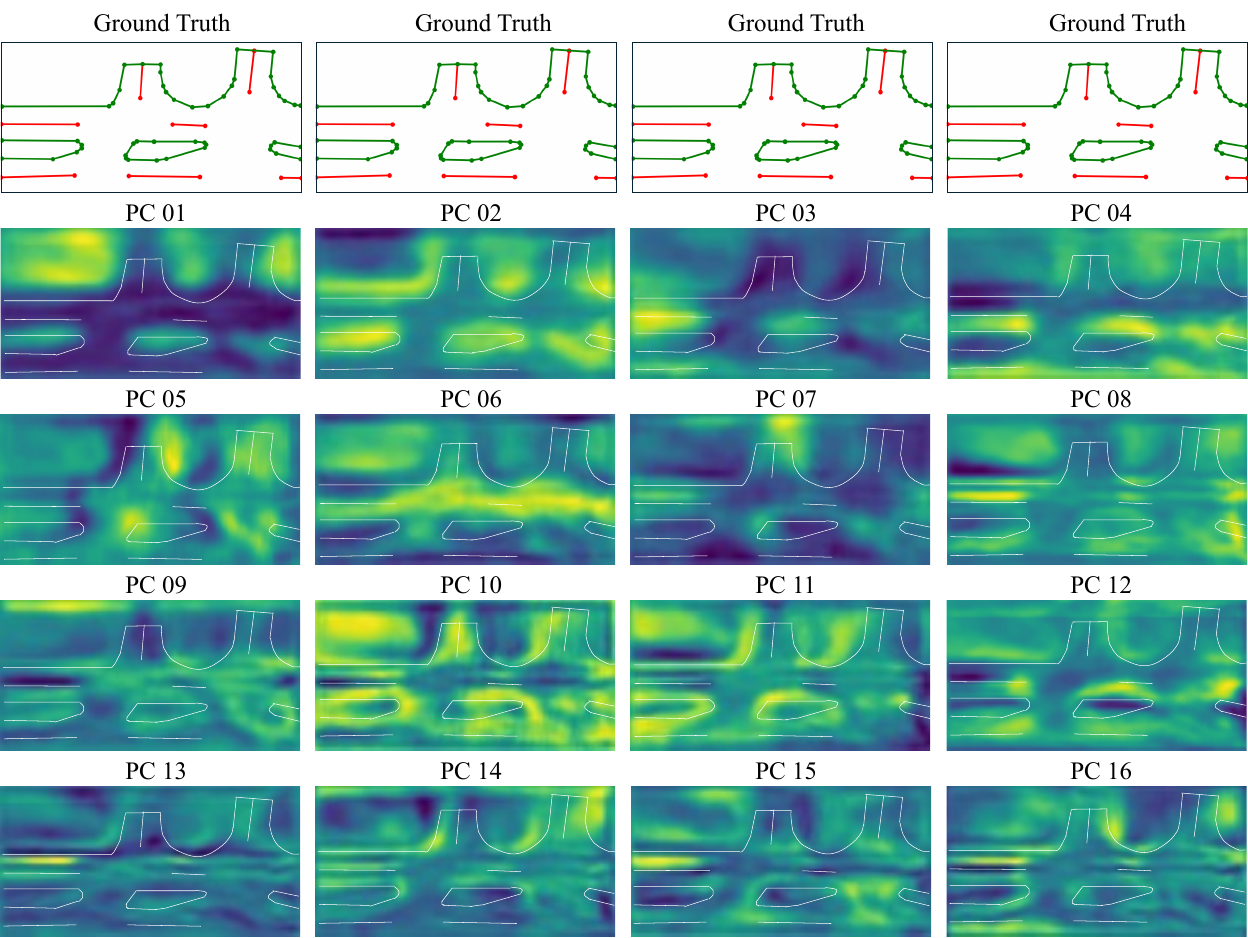}
        \caption{AugMapNet}
        \label{fig:top16_pc_amn_ex2}
    \end{subfigure}
   \caption{Example scene 2: Top 16 Principal Components of latent BEV grid of (a) StreamMapNet and (b) AugMapNet. Vector map ground truth is visualized at the top of each column as well as overlaid as white lines to ease the assessment of spatial accuracy. Notice that StreamMapNet tends to have radial artifacts.}
   \label{fig:top16_pc_ex2}
\end{figure*}

As an extension to the top 3 Principal Components (PCs) visualized in \cref{fig:latent_pca}, we visualize the top 16 PCs for example scene 1 in \cref{fig:top16_pc_ex1} and for example scene 2 in \cref{fig:top16_pc_ex2}.
A general observation is that the PCs tend to have more radial artifacts in StreamMapNet.
The PCs of AugMapNet have better spatial structure and correspondence with the GT map due to dense spatial supervision.

\begin{figure*}
        \centering
        \includegraphics[width=0.6\textwidth]{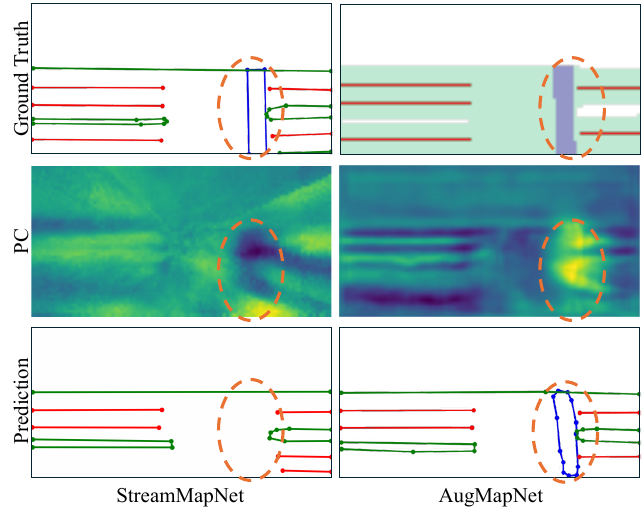}
   \caption{Extension of \cref{fig:latent_pca}, visualizing example scene 1 where pedestrian crossing is missed by StreamMapNet but predicted by AugMapNet. PC is the Principal Component of the BEV latent grid that has the highest visual correspondence to the pedestrian crossing out of the top 16 Principal Components visualized in \cref{fig:top16_pc_ex1}. It is PC 14 for StreamMapNet and PC 8 for AugMapNet, suggesting that this \enquote{crossing} PC is stronger in AugMapNet than StreamMapNet.}
   \label{fig:pc_activation}
\end{figure*}

Finally, \cref{fig:pc_activation} gives more details on example scene 1, where StreamMapNet misses a pedestrian crossing that is correctly predicted by AugMapNet.
Specifically, we highlight the PC that has the highest visual correspondence to the pedestrian crossing GT out of the top 16 PCs visualized in \cref{fig:top16_pc_ex1} for each StreamMapNet and AugMapNet.
It is PC 14 for StreamMapNet and PC 8 for AugMapNet.
The smaller number for AugMapNet indicates that AugMapNet has stronger latent features for pedestrian crossings, which likely helped with its correct prediction.
This also matches the better visual correspondence between the AugMapNet PC and the pedestrian crossing GT compared to StreamMapNet.

\clearpage

\end{document}